\documentclass{article}
\usepackage{amsmath}
\usepackage{graphicx}
\usepackage[dvipsnames,table]{xcolor}
\usepackage{fontawesome}
\usepackage{enumitem}
\usepackage{soul}
\usepackage[ruled, linesnumbered]{algorithm2e}

 \usepackage[main, final]{neurips_2026}

\usepackage[utf8]{inputenc} 
\usepackage[T1]{fontenc}    
\usepackage{hyperref}       
\usepackage{url}            
\usepackage{booktabs}       
\usepackage{amsmath}
\usepackage{amssymb}
\usepackage{amsfonts}       
\usepackage{nicefrac}       
\usepackage{microtype}      
\usepackage{xcolor}         
\usepackage{makecell}

\newcommand{\errbar}[1]{{\footnotesize\,\color{gray}$\pm$#1}}
\newcommand{\cinterval}[2]{{\footnotesize~\color{gray}(#1,\,#2)}}

\newcommand{\expensive}{\,{\tiny\color{Red!75!Black}\raisebox{0.8pt}{\faHourglass[3]}}}
\newcommand{\mycomment}[1]{}
\title{Conveyance: A Versatile Framework for Learning in Structured Class Spaces}

\author{%
  Yasser Taha$^{1,2}$
  \qquad
  Grégoire Montavon$^{2,3,}$\thanks{Equal supervision.}
  \qquad
  Nils Körber$^{1,}$\footnotemark[1]
  \\[1ex]
  {\normalfont\small
  \begin{tabular}{@{}c@{}}
  $^{1}$Centre for Artificial Intelligence in Public Health Research (ZKI-PH), Robert Koch Institute\\
  $^{2}$Institute for Artificial Intelligence in Medicine (IKIM), Charité -- Universitätsmedizin Berlin\\
  $^{3}$BIFOLD -- Berlin Institute for the Foundations of Learning and Data\\[0.8ex]
  \texttt{\{tahay,\,koerbern\}@rki.de}
  \qquad
  \texttt{gregoire.montavon@charite.de}
  \end{tabular}}
}

\makeatletter
\let\conv@maketitle\@maketitle
\renewcommand{\@maketitle}{%
  \let\conv@title\@title
  \def\@title{\hrule height 4pt\vskip 0.25in
    \conv@title\par\vskip 0.29in\hrule height 1pt}%
  \conv@maketitle}
\makeatother
\begin{document}

\maketitle

\begin{abstract}
While machine learning (ML) architectures have evolved rapidly to account for complex data, loss functions like cross-entropy remain mostly structure-agnostic in many real-world applications.
However, the `class-symmetric' nature of these standard losses fundamentally limits the ability of ML models to exploit structural relationships between classes, particularly when facing structured noise.
We propose \textsc{Conveyance}, a new classification approach and associated loss function tailored to structured class spaces. It allows users to encode graph-like relations between classes without having to define complex joint distributions or manually tune utility matrices.
Technically, our loss function operates by maximizing two separate margins over distinct class partitions, while preserving formal properties such as monotonicity and partial convexity.
We demonstrate the versatility and effectiveness of our method by applying it to hierarchical classification, ordinal regression, and multiple instance learning. Across these tasks, \textsc{Conveyance} either matches or exceeds the performance of specialized baselines, thereby offering a unified solution for structured class spaces.
\end{abstract}

\section{Introduction}
\label{sec:intro}
Despite the rapid evolution of model architectures within the deep learning landscape, training objectives have remained remarkably consistent. Loss functions such as cross-entropy continue to be workhorses of the field, widely used in both research and applications, and the default across major software frameworks.
These classification losses assign a scalar penalty to each prediction based solely on whether it matches the target label, treating all errors as interchangeable. This structure-agnostic assumption is adequate when class boundaries are sharp and labels are clean, but it becomes a fundamental liability in two broad and practically important regimes.

The first is \textbf{labeling asymmetry}. These asymmetries arise in several contexts, including labeler bias, underdiagnosis from partial observability in medical imaging, and taxonomic shifts such as the WHO's periodic ICD code revisions. Another prominent example occurs when framing Multiple Instance Learning (MIL) as patch-wise classification: often, the majority of constituent patches are semantically negative yet inherit the bag's positive label. Treating every local instance as a standard categorical target would thus impose a systematically erroneous supervision signal.
The second is \textbf{semantic and ordinal continuity}.  In many domains, categories are discrete proxies for an underlying continuum: a related species within the same genus, adjacent stages of a progressive disease, or neighboring age groups are not arbitrary as they occupy geometrically close positions in input space.  A near-miss in these settings is semantically far less costly than a random misclassification, yet cross-entropy (CE) assigns both the same unit penalty.  The consequence is well documented: CE-trained models overfit to target labels, forcing separation between neighboring manifold points and fragmenting the latent space in ways that degrade generalization and calibration.

Existing remedies address these regimes in isolation, each carrying its own cost.  Soft-label methods~\cite{label_smoothing,soft_labels,dldl} replace one-hot targets with fixed parametric distributions (e.g. uniform, Laplace, Gaussian), but their symmetric, pre-specified forms cannot capture asymmetric or graph-structured relations. Rank-consistent ordinal methods such as CORAL~\cite{coral} and CORN~\cite{corn} instead replace the softmax head with a set of ordered binary classifiers; this guarantees consistent rank predictions, but requires changing the output layer and applies only to linearly ordered labels. Structured output SVMs~\cite{ssvm} and energy-based models~\cite{ebl} provide principled cost-augmented margins, yet require inference over an exponentially large output space via non-differentiable algorithms, precluding end-to-end learning.  Contrastive losses~\cite{triplet_loss,supcon} cluster related classes in embedding space at the price of large batches and sensitive pair mining.  Noise transition methods~\cite{masking} restrict invalid label mappings through a binary matrix but still rely on generative noise estimation, adding substantial computational overhead.  No existing framework is simultaneously structurally expressive, discriminative, and as computationally simple as CE.

We propose \textsc{Conveyance}, a novel approach to learning classifiers that consists of two main components: (i) a \ul{knowledge encoding scheme} for class space structure, utilizing an easy-to-specify Boolean matrix $Q$ to link labels to plausible classes; and (ii) a \ul{loss function}, which operationalizes this scheme through \textit{two} mutually dependent margins representing predicted evidence for the plausible set and the assigned label, respectively. Our loss requires no architectural changes, sampling procedures, or generative models. While it recovers standard CE as a limit case, it offers significantly greater flexibility in handling structured class spaces and label noise. We demonstrate this versatility through (iii) \ul{three practical instantiations} of our approach: label asymmetry (synthetic label noise and multiple instance learning), ordinal regression (age estimation), and hierarchical classification (taxonomic classification). In each case, \textsc{Conveyance} matches or exceeds specialized state-of-the-art methods. (i)--(iii) constitute our three main contributions. Our code is available at \url{https://github.com/ZKI-PH-ImageAnalysis/Conveyance}.

\begin{figure}[t!]
    \centering
    \includegraphics[width=\linewidth]{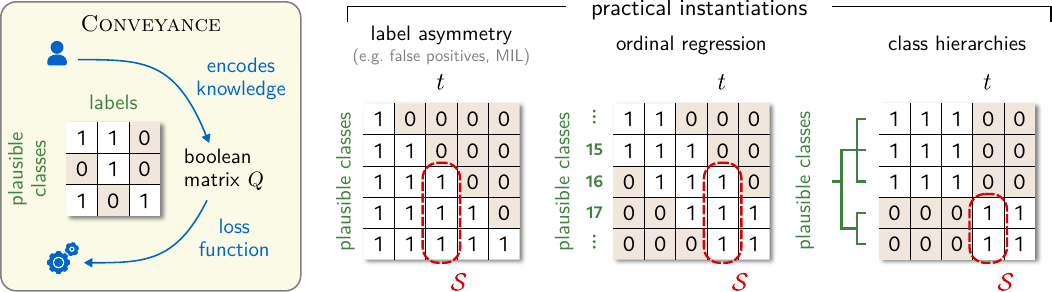}
    \caption{Overview of the proposed \textsc{Conveyance} method for learning in structured class spaces. The method encodes problem knowledge through a boolean matrix $Q$ connecting each label $t$ to a set of plausible classes $\mathcal{S}$, and feeds this information into a purposely designed loss function (Eq.\ \eqref{eq:conveyance}). The approach is versatile, expressing a variety of supervised learning problems such as learning under label asymmetry, ordinal regression, and hierarchical classification (from left to right).}
    \label{fig:placeholder}
\end{figure}

\section{Related work}

\textbf{Fixed priors and label smoothing.}
To incorporate class relationships, several methods optimize the model for `soft targets': Extending the principle of label smoothing~\cite{label_smoothing} which spreads a uniform residual mass over all classes, methods such as Soft Labels~\cite{soft_labels} and DLDL~\cite{dldl} soften targets using ordinal priors (e.g.\ Laplace or Gaussian decays centered on the original target). Unimodal \cite{unimodal} extends this line of work by proposing unimodal parametric families (binomial, Poisson) as ordinal targets, obtaining a differentiable objective that respects rank order. A complementary direction bypasses parametric families to directly optimize the Earth Mover's Distance (EMD) between the predicted distribution and a one-hot or soft target~\cite{emd}. Despite their differences, those methods typically impose a pre-specified distributional prior over linearly ordered labels, which prevents them from capturing asymmetric relations, hierarchical taxonomies, or graph-structured dependencies. \textsc{Conveyance} avoids any distributional assumption, maximizing instead margins over user-defined class partitions.

\textbf{Class-label transition matrices.}
A step towards more expressiveness is taken by the method of \cite{Patrini2017} that assumes a noisy (potentially asymmetric) transition model from classes to labels, and reverses it through backward/forward correction mechanisms built into the architecture, assuming full knowledge of the underlying transition probabilities. Concurrent work attempts to learn the transition matrix end-to-end without pre-estimation~\cite{adaptation}, though brute-force optimization over the full matrix can suffer from local minima with finite data. Later methods address this instability through more constrained parameterizations~\cite{dirichlet}, while \cite{variation} further avoids explicit posterior estimation by regularizing predicted probabilities to be maximally distinguishable. Closer to our approach, Masking~\cite{masking} relaxes the requirement for a dense transition matrix, by requiring only a binary matrix encoding which class transitions are plausible (similar to $Q$ in our paper), using it as a structural prior to constrain the estimation of underlying noise transition probabilities. In contrast, \textsc{Conveyance} avoids estimating transition probabilities: it operationalizes $Q$ directly as margin constraints in logit space, bypassing probabilistic modeling entirely at a computational cost identical to standard cross-entropy. 

\textbf{Partial-label and set-valued losses.}
Partial-label learning~\cite{cour2011learning,feng2020provably} addresses a related setting in which the true label is unknown and each sample only comes with a set of candidate labels. In \textsc{Conveyance}, by contrast, the label is observed and kept as a target, and the plausible set encodes prior structural knowledge about the class space rather than annotation ambiguity.

\textbf{Energy based learning and structured margins.}
Beyond classical class-label models, our approach draws inspiration from Energy-Based Learning (EBL)~\cite{ebl} and Structured Output SVMs (SSVMs)~\cite{ssvm}, which model class dependencies through cost-augmented margins. In SSVMs, the objective is to separate the ground-truth $y^\star$ structure from competing hypotheses $y$ by a margin lower-bounded by a task-specific cost function. While highly effective for structured prediction, classic SSVMs require inference over a combinatorially large output space (e.g, all graph configurations or label sequences), often relying on complex, non-differentiable inference algorithms such as Viterbi~\cite{viterbi} or Graph Cuts~\cite{graphcut}. In contrast, \textsc{Conveyance} operates directly on top of the logits, which endows it with computational efficiency (on par with cross-entropy) and integrability in a wide range of gradient-based ML pipelines.

\textbf{Structured representation learning.} Another body of related work addresses label relationships through representation learning. Methods such as Triplet Loss~\cite{triplet_loss} and Supervised Contrastive (SupCon) learning~\cite{supcon} shape the embedding space via pairwise supervision to cluster semantically related samples. While effective for representation learning and also relying technically on maximizing margins, these approaches operate on inter-sample relationships in high-dimensional embedding space. By leveraging label topology encoded in a boolean matrix $Q$ rather than on a distribution of sample pairs, \textsc{Conveyance} more directly addresses the model's classification behavior.

\section{Proposed Method}
\label{sec:method}
We introduce \textsc{Conveyance}, a versatile approach for learning classifiers in structured class spaces and to mitigate structured label noise. For simplicity, we assume that the label and the actual class share the same space, denoted as $\mathcal{C} = \{1,\dots,C\}$. Our method starts by having the user specify a boolean matrix $Q \in \{0,1\}^{C \times C}$ where each element $Q_{ct}$ indicates whether it is conceivable that an instance labeled as $t$ has actual class $c$, or conversely, that an instance of class $c$ is labeled as $t$. Positive off-diagonal terms may encode the plausibility of false positives or negatives, the risk of mixing between two specializations of a common class, or of mixing between two nearby ordinals (e.g.\ age). We note that such a specification is much more tractable for the user than specifying a full joint probability distribution over classes and labels, and also more robust to shifts in the data. In many applications, $Q$ is dictated by the problem itself: it follows from a known corruption mechanism, a taxonomy, or ordinal adjacency. Where it is only approximately known, our robustness study (Sec.~\ref{sec:q-robustness}) shows that over-specifying plausible transitions is safer than omitting true ones.

\textbf{Conveyance loss.} We now translate the user's specification (the matrix $Q$) into an optimizable loss function. We construct the set $\mathcal{S} = \{c \in \mathcal{C}  : Q_{ct} = 1\}$ containing all plausible classes given the label is $t$. We assume the classifier outputs a vector $(p_c)_{c \in \mathcal{C}}$ containing class probabilities, and denote by $p_\mathcal{S} = \sum_{s \in \mathcal{S}} p_s$ the total probability of plausible classes. We define the Conveyance loss as:
\begin{align}
 \ell(p,t) &= \log \Big(
1 +
\alpha \cdot \frac{1-p_t}{p_t} +
\beta \cdot \frac{1-p_\mathcal{S}}{p_\mathcal{S}}
 \Big)
 \label{eq:conveyance}
\end{align}
The first ratio rewards allocating probability to the target class over the non-target classes. The second ratio rewards allocating probability to the plausible set over the non-plausible set. When setting hyperparameters to $\alpha = 1$ and $\beta=0$ the loss reduces to cross-entropy. 
The desired Conveyance loss behavior is achieved by reducing $\alpha$ and increasing $\beta$ (e.g., $\alpha=0.1,\beta=10$), making allocation to $\mathcal{S}$ the primary modeling force, and relegating $t$ to a tie-breaker.

\begin{figure}[h]
    \centering
    \includegraphics[width=\linewidth]{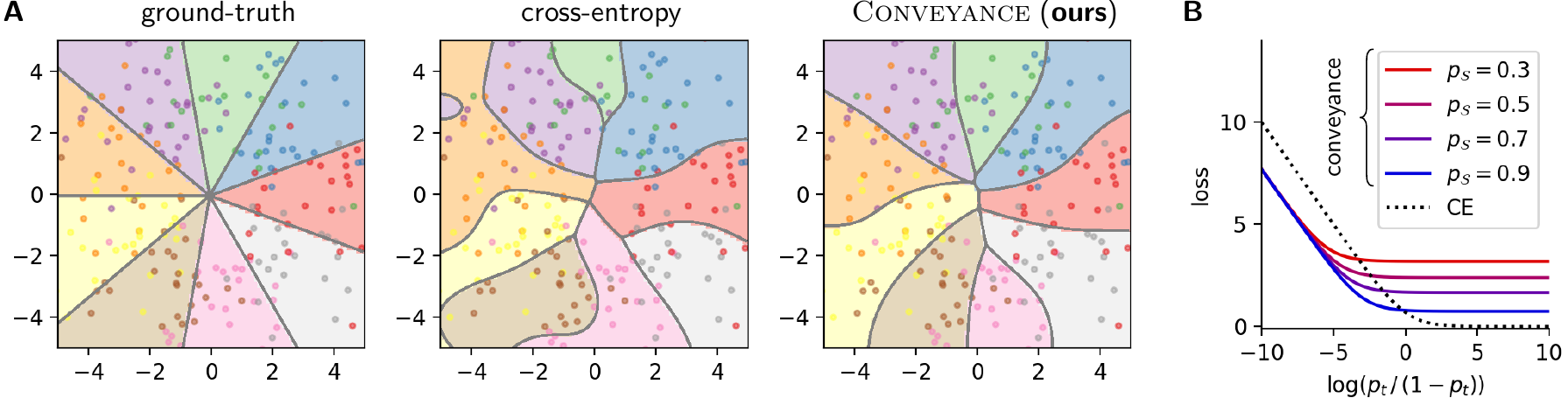}
    \caption{\textbf{\sffamily \small A.} Application of \textsc{Conveyance} on a noisy two-dimensional problem where angles define the true class, and where we set $\mathcal{S} = \{t-1,t,t+1\}$. This encoded knowledge, together with the double-margin characteristic of the loss function, helps produce robust decision boundaries. \textbf{\sffamily \small B.} Monotonicity of the Conveyance loss (from left to right, and from red to blue). The loss resembles cross-entropy (CE), but $p_\mathcal{S}$ modulates the geometry of the $p_t$-margin. (Parameters: $\alpha=0.1, \beta=10$.)}
    \label{fig:toy}
\end{figure}

\textbf{Theoretical analysis.} As a starting point for our analysis, we consider the same loss as in Eq.\ \eqref{eq:conveyance} but rewritten as a log-sum-exp (LSE):
\begin{align}
 \ell(p,t) &= \mathrm{LSE} \Big\{
 0 \,,\,
 \log\Big(\frac{1-p_t}{p_t}\Big) + \log \alpha \,,\,
 \log \Big(\frac{1-p_\mathcal{S}}{p_\mathcal{S}}\Big) + \log \beta 
\Big\}
 \label{eq:conveyance-expanded}
\end{align}
(Derivation in Appendix \ref{sec:derivation}). The LSE operation can be interpreted as a soft max-pooling, here applying to a zero-valued constant and two log-odds. It effectively implements an advanced \ul{margin loss}, where not one, but \textit{two} margins need to be achieved simultaneously: separating target from non-target, and separating plausible from implausible classes. The qualitative effect of this advanced margin formulation can be observed on a simple two-dimensional example in Fig.\ \ref{fig:toy} (left). In this formulation, the hyperparameters $\alpha$ and $\beta$ can be interpreted as placing priors on which margin is active in the pool. The form of Eq.\ \ref{eq:conveyance-expanded} also allows us to study \textsc{Conveyance}'s \ul{monotonicity}: Observing that LSE and log-odds are monotonic with their arguments, we can show that any reallocation of probability $p \to p^\prime$ that increases both $p_t$ and $p_\mathcal{S}$ causes the Conveyance loss to decrease. This desirable monotonic behavior can also be observed in Fig.\ \ref{fig:toy} (right).

We now assume that probabilities are given by a softmax layer $p_c = \exp(z_c)/\mathcal{Z}$, with $z_c$ the real-valued logits. We can show that the Conveyance Loss has \ul{partial convexity} with $z$, specifically, treating $\{z_s\}_{s \in \mathcal{S}}$ as constants, the loss becomes convex in $\{z_n\}_{n \in \mathcal{N}}$ where $\mathcal{N} = \mathcal{C}-\mathcal{S}$. To demonstrate this, we further develop Eq.\ \eqref{eq:conveyance-expanded} as:
\begin{align}
\ell(p,t) = \mathrm{LSE} \Big\{ 0
\,,\,
\big(z_c - z_t + \log \alpha\big)_{c \in \mathcal{C} - \{t\}}
\,,\,
\big(z_n -  \mathrm{LSE}_{s \in \mathcal{S}} \{z_s\} + \log \beta\big)_{n \in \mathcal{N}}
 \Big\}
 \label{eq:conveyance-more}
\end{align}
(Derivation in Appendix \ref{sec:derivation}). Observing that LSE is a convex function, and that it is composed with affine functions of $\{z_n\}_{n \in \mathcal{N}}$ demonstrates the property. This convexity result suggests reasonable optimization performance when $\mathcal{S}$ is small and varying across the dataset. For the limit case where $|\mathcal{S}| = 1$, the function to optimize becomes again fully convex. Additionally, the form of Eq.\ \eqref{eq:conveyance-more} is also more stable numerically, and thus constitutes a natural starting point for implementation (cf.\ Appendix \ref{sec:pseudo}).

\begin{figure}
    \centering
    \includegraphics[width=\linewidth]{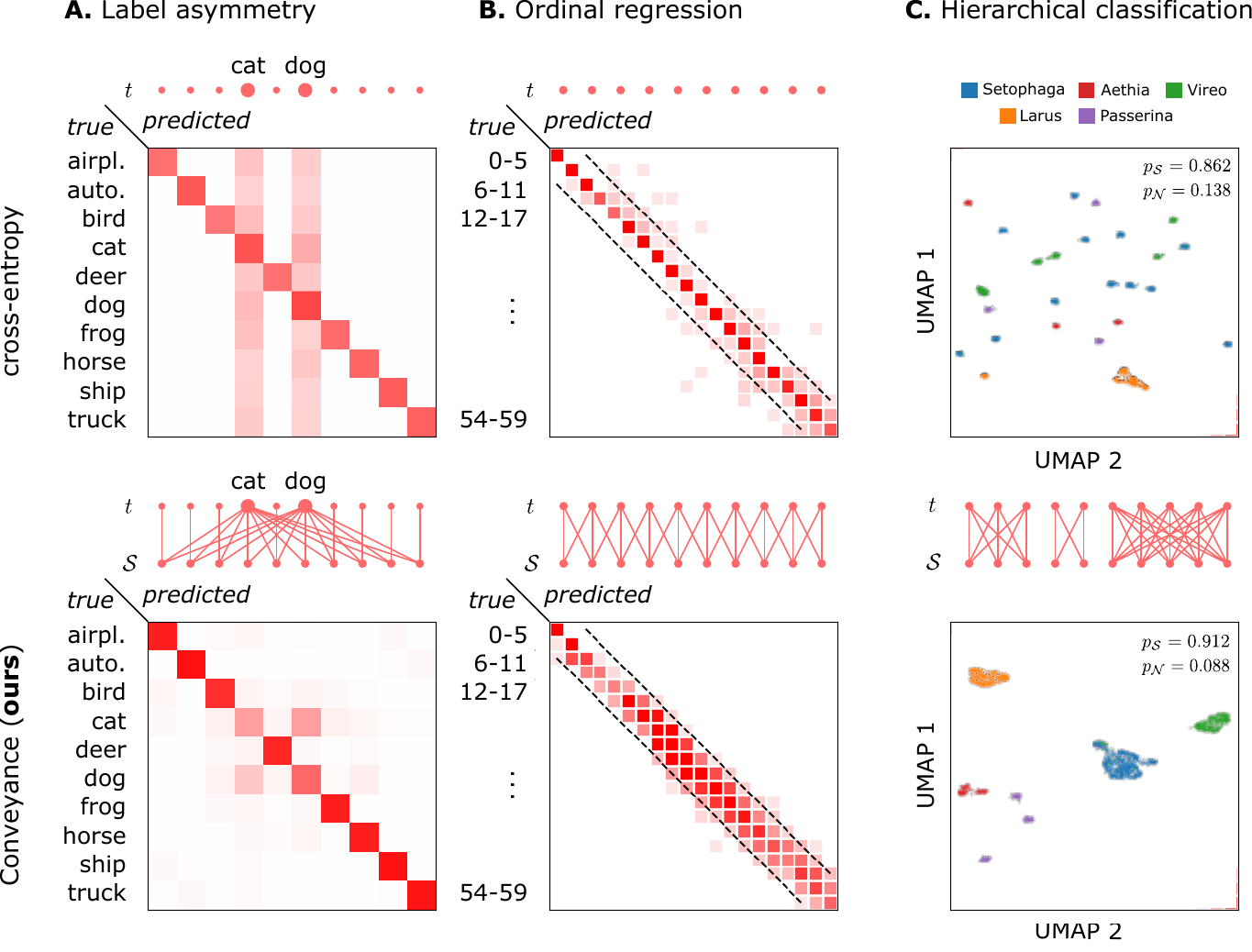}
  \caption{%
  Overview of \textsc{Conveyance}'s qualitative behavior across three structured classification tasks.
  \textbf{\sffamily \small A.} Synthetic setting, where $\eta = 60\%$ of CIFAR-10 instances are wrongly annotated to $t$=`cat' or $t$=`dog'. Predictions of the noise-agnostic CE baseline are strongly biased towards these two dominant classes. Our method, together with its encoded knowledge $t \to \mathcal{S}$, is able to recover a near-diagonal structure.
  \textbf{\sffamily \small B.} True vs.\ predicted age on CLAP-2016 training data. CE predictions spread laterally off the diagonal at the tails, whereas our method concentrates density along the diagonal for the full age range.
  \textbf{\sffamily \small C.} UMAP of feature embeddings on the CUB-200 bird classification data colored by genus (five most frequent genera). CE fragments species into overlapping clouds, whereas our method forms fewer, larger clusters reflecting the latent taxonomic structure. A similar UMAP for the ordinal regression application can be found in Appendix \ref{sec:analysis-app}.
}
    \label{fig:results}
\end{figure}

\section{Experiments}
\label{sec:experiments}

We evaluate the performance and versatility of \textsc{Conveyance} across three distinct domains: label asymmetry (annotation bias and multiple instance learning), ordinal regression and hierarchical classification. For each application, we benchmark our method against SOTA methods, maintaining identical architectures and training protocols to isolate the impact of the objective function. A qualitative overview of our method's behavior on all three tasks is provided in Fig.\ \ref{fig:results}. The detailed experimental procedures and descriptions of all 12 benchmark datasets are provided in Appendix ~\ref{sec:exp-details}.


\subsection{Label asymmetry}
Label asymmetry arises whenever the relationship between observed labels and true classes is structurally non-uniform: some transitions are possible, others are not, and the direction matters.  We study this regime in two complementary settings.

\textbf{Annotation bias:} We evaluate \textsc{Conveyance} against the generative \textit{Masking} framework \cite{masking} across four distinct configurations: On CIFAR-10, Column noise ($60\%$) and Asymmetric noise ($45\%$). On CIFAR-100, Block noise ($60\%$) and Asymmetric noise ($45\%$). Detailed explanation of each noise type is deferred to Appendix ~\ref{sec:noise-types}.

Table~\ref{tab:masking} reports top-1 test accuracy and TFLOPS for CE, Masking~\cite{masking}, and \textsc{Conveyance} ($\alpha{=}0.1, \beta{=}10$) on CIFAR-10 and CIFAR-100 under annotation bias. Our method ranks first across all four noise settings and is computationally much cheaper than Masking. Under asymmetric noise, it closely approaches the accuracy of a CE model trained on clean data. This highlights our method's near-perfect denoising capability.
In fact, on CIFAR-100, with different $\alpha,\beta$ settings, it \emph{surpasses} the clean baseline by $+3.56$ pp ($73.70\%$ vs.\ $70.14\%$), suggesting that source-set grouping acts as an implicit structural regularizer that suppresses spurious inter-class confidence beyond what a noise-free signal alone provides (see Section~\ref{sec:ablation-main}).
\textsc{Conveyance} receives only a \emph{binary} transition mask $Q$, without true noise rates, yet learns to concentrate probability mass on plausible classes in proportion to the true underlying transition frequencies, effectively recovering the true noise rates from the training signal alone. This emergent recovery is visible in Figure~\ref{fig:results}: under column noise ($\eta{=}0.6$), CE collapses toward the dominant annotations cat and dog, while \textsc{Conveyance} recovers a near-diagonal prediction structure with residual confusion confined to the cat/dog pair, closely matching the ground-truth transition structure that CE cannot reproduce.
\begin{table}[t!]
\caption{\textbf{Left:} Test accuracy (\%) and TFLOPS on CIFAR~\cite{cifar10-100} under structured label noise. Mean\,$\pm$\,std are reported over 3 seeds. ($^\ast$) results copied from the original paper. \textbf{Right:}Validation accuracy (\%) of \textsc{Conveyance} across the $\alpha$--$\beta$ parameter grid on CIFAR-10 (top) and CIFAR-100 (bottom) under structured noise. Full numerical heatmaps are provided in Appendix~\ref{sec:ablations}.}
  \label{tab:masking}
  \begin{minipage}{.55\textwidth}
  \small
  \addtolength{\tabcolsep}{-0.4em}
  \begin{tabular}{lccc}
  \toprule
  Method
    & \makecell{asymmetric
    } 
    & \makecell{col./block
    }
    & \!TFLOPS\!\\
  \midrule
  \multicolumn{3}{l}{\itshape CIFAR-10 (clean: 91.67\errbar{0.42})}\\[1mm]
  ~~CE        & 80.53\errbar{0.61}    & 54.70\errbar{0.72} &0.8\\
  ~~Masking~\cite{masking}  & 83.54\errbar{3.21} & 75.42\errbar{2.84}& 15.4\expensive{}\\
  \rowcolor{cyan!10} ~~Conveyance (\textbf{ours}) & \textbf{89.68}\errbar{0.21}    & \textbf{80.02}\errbar{0.56} &0.8\\~\\[-1mm]
  \multicolumn{3}{l}{\itshape CIFAR-100 (clean: 70.14\errbar{0.25})}\\[1mm]
  ~~CE        & 41.03\errbar{0.61}& 56.87\errbar{0.92}  &11.5\\
  ~~Masking~\cite{masking} &56.72\errbar{1.54}&56.50$^\ast$\phantom{\errbar{0.0\,}} &134\expensive{} \\
  \rowcolor{cyan!10} ~~Conveyance (\textbf{ours}) & \textbf{70.43}\errbar{0.44}& \textbf{63.21}\errbar{0.85} &11.5 \\
  \bottomrule
\end{tabular}
\end{minipage}
\hfill
\begin{minipage}{.43\textwidth} \centering
    \includegraphics[width=\linewidth,height=5.5cm,keepaspectratio]{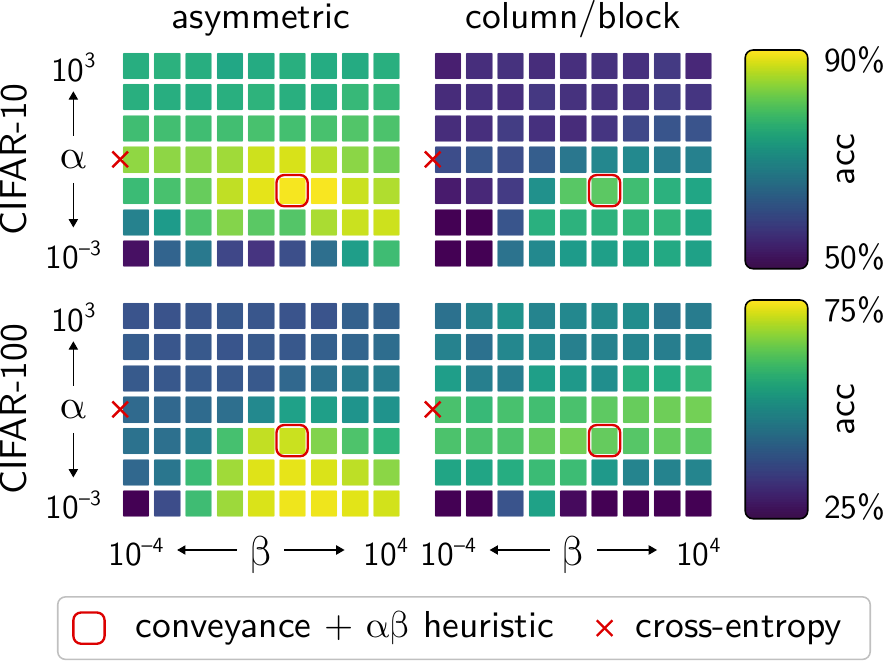}
\end{minipage}
\end{table}

\ul{Ablation\,/\,sensitivity analysis.}
\label{sec:ablation-main}
To verify that \textsc{Conveyance} benefits from its two parameters $\alpha, \beta$ and to test how predictable the relation between parameter, task, and model performance is, we sweep $\alpha, \beta \in \{10^{-4}, \ldots, 10^{4}\}$ across all datasets and noise settings considered in Table \ref{tab:masking} and display validation accuracy as heatmaps to the right of that table. (Detailed heatmaps and further ablations are in Appendices~\ref{sec:noise-types} and \ref{sec:ablations}.) We observe that $\alpha$ and $\beta$ play structurally distinct and complementary roles determined by the noise topology. $\beta$ penalizes probability mass \emph{outside} the source-augmented set $\mathcal{S}$ and is therefore active only when $p_N{>}0$, while $\alpha$ governs the odds term and acts on every sample unconditionally. For both datasets and all noise settings, we observe optimal values near $(\alpha{=}0.1,\, \beta{=}10)$, indicating a tradeoff between a small $\alpha$ to avoid hardening noisy labels and a moderate $\beta$ to refine source-class confidence. However, the sharpness of the peak is dependent on the noise setup. In our experiments, we found the relevant range to be $\alpha \in [0.1, 1]$ and $\beta \in [1, 10]$, making them easy-to-tune hyperparameters. 

\ul{Robustness to a misspecified $Q$.}
\label{sec:q-robustness}
So far, $Q$ was set to the support of the corruption process ($Q_{ct}{=}1$ iff $T_{ct}{>}0$; Appendix~\ref{sec:noise-types}), which might not be known with certainty in practice. We therefore keep the label corruption on CIFAR-10 fixed and distort only the $Q$ given to the loss, in two ways: a \emph{broad} $Q$ adds spurious off-diagonal entries (over-specification), while a \emph{conservative} $Q$ removes true ones (under-specification). Table~\ref{tab:q-robustness} shows that the two failure modes are not symmetric. Over-specification degrades gracefully: under asymmetric noise, doubling the number of off-diagonal entries costs only 2.1\,pp and still leaves \textsc{Conveyance} 7.1\,pp above CE, and under column noise a $Q$ with additional 50\% spurious entries retains nearly half of the gain over CE. Under-specification is more harmful but fails safely: as true entries are removed, $\mathcal{S}$ shrinks towards $\{t\}$, for which the loss becomes the CE-like objective $\log(1+(\alpha{+}\beta)(1-p_t)/p_t)$, and accuracy falls back to the CE level. Column noise, which funnels all corruption into two classes, is the more sensitive of the two settings. In practice, these results suggest including a transition in $Q$ whenever in doubt.

\begin{table}[t]
\caption{Robustness to a misspecified $Q$ on CIFAR-10: test accuracy (\%, mean\,$\pm$\,std over 3 seeds). The label corruption is unchanged; only the $Q$ given to the loss is distorted. \emph{Broad}: spurious off-diagonal entries added; \emph{conservative}: true off-diagonal entries removed. Misspecification is measured relative to the number of true off-diagonal entries in $Q$ (18 under column noise, 4 under asymmetric noise). CE is the structure-agnostic reference from Table~\ref{tab:masking}.}
\label{tab:q-robustness}
\centering
\small
\begin{minipage}[t]{.48\linewidth}
\centering
\textit{Column noise} ($\eta{=}0.6$)\\[1mm]
\begin{tabular}{lcc}
\toprule
Misspec. & Broad $Q$ & Conservative $Q$ \\
\midrule
\textit{CE} & \multicolumn{2}{c}{\textit{54.7}\errbar{0.7}} \\
0\% (true $Q$) & \multicolumn{2}{c}{\textbf{80.0}\errbar{0.6}} \\
10\% & 78.0\errbar{0.8} & 73.8\errbar{0.4} \\
50\% & 66.3\errbar{4.9} & 55.4\errbar{1.4} \\
100\% & 57.9\errbar{5.3} & 54.8\errbar{0.6} \\
\bottomrule
\end{tabular}
\end{minipage}
\hfill
\begin{minipage}[t]{.48\linewidth}
\centering
\textit{Asymmetric noise} ($\eta{=}0.45$)\\[1mm]
\begin{tabular}{lcc}
\toprule
Misspec. & Broad $Q$ & Conservative $Q$ \\
\midrule
\textit{CE} & \multicolumn{2}{c}{\textit{80.5}\errbar{0.6}} \\
0\% (true $Q$) & \multicolumn{2}{c}{\textbf{89.7}\errbar{0.2}} \\
25\% & 89.3\errbar{0.3} & 86.6\errbar{0.5} \\
50\% & 88.6\errbar{0.3} & 84.5\errbar{0.8} \\
75\% & 87.7\errbar{2.2} & 83.1\errbar{0.9} \\
100\% & 87.6\errbar{1.6} & 80.4\errbar{0.2} \\
\bottomrule
\end{tabular}
\end{minipage}
\end{table}




\textbf{Multiple instance learning (MIL):} 
We evaluate \textsc{Conveyance} across two distinct MIL regimes: the Camelyon-16~\cite{camelyon_16} WSI benchmark and five classic tabular datasets~\cite{musks,animals}. MIL is characterized by a fundamental label asymmetry: while a negative bag contains strictly negative instances, a positive bag may contain a mix of positive and "false positive" instances. We encode this structural constraint directly into our binary transition matrix $Q$, where $Q_{\text{neg}\to\text{pos}}{=}1$ and $Q_{\text{pos}\to\text{neg}}{=}0$ (see Fig.\ \ref{fig:placeholder}). Following standard practice, we employ a Bag-level Cross-Entropy loss; however, we augment this with \textsc{Conveyance} at the instance level to provide a structured inductive bias. For a fair comparison, we utilize the identical architectures, data splits, and training protocols as the state-of-the-art baselines reported in~\cite{psamil} (see Appendix~\ref{sec:implementation-details} for further details).

Table~\ref{tab:mil-benchmarks} reports results on five classic MIL benchmarks (10-fold cross-validation (CV), 5 runs).  \textsc{Conveyance} achieves the highest accuracy on ELEPHANT and places second on MUSK2, FOX and TIGER, where confidence intervals largely overlap with the best-performing baselines.  It matches or surpasses specialized methods such as TR-RGMIL~\cite{tr_rgmil} and PSMIL~\cite{psamil} on four of five datasets despite using no problem-specific design choices.
Table~\ref{tab:camelyon16} reports results on the Camelyon-16 whole-slide image benchmark. \textsc{Conveyance} achieves state-of-the-art accuracy ($0.933$) and AUC ($0.977$), notably improving AUC by $1.8$ points over the next-best method, CAMIL ($0.959$), with non-overlapping confidence intervals. While F1 scores remain competitive, they are slightly lower than DTFD-MIL~\cite{dtfdmil} and PSMIL~\cite{psamil}. This margin is expected, as these baselines rely on architectural specializations, specifically, DTFD-MIL's multi-tier feature distillation and PSMIL's prototype-based latent clustering, whereas \textsc{Conveyance} achieves superior discriminative power solely through its structured objective function.

\begin{table}[htbp]
  \caption{Accuracy on five classic tabular MIL benchmarks (10-fold CV, 5 runs; mean\,$\pm$\,std). \textbf{Bold}: best, \underline{underline}: second best. Conveyance, applied with default parameters and no MIL-specific design choices.}
  \label{tab:mil-benchmarks}
  \centering \small
  \begin{tabular}{lccccc}
    \toprule

  Method & MUSK1\,($\uparrow$) & MUSK2\,($\uparrow$)& FOX\,($\uparrow$)& TIGER\,($\uparrow$) & ELEPHANT\,($\uparrow$)\\
    \midrule
    ABMIL~\cite{abmil} &
    0.892\errbar{0.040}  &  0.858\errbar{0.048}  &  0.615\errbar{0.043} & 0.839\errbar{0.022}& 0.868\errbar{0.022}\\
    Gated-ABMIL~\cite{abmil}  &0.900\errbar{0.050}  &  0.863\errbar{0.042}  &  0.603\errbar{0.029} & 0.845\errbar{0.018}& 0.857\errbar{0.027}\\
    DPMIL~\cite{dpmil}  &0.907\errbar{0.036}  &  0.926\errbar{0.043}  &  0.655\errbar{0.052} & 0.897\errbar{0.028}& 0.894\errbar{0.030}\\
    DSMIL~\cite{dsmil}  &0.932\errbar{0.023}  &  0.930\errbar{0.020}  &  0.729\errbar{0.018} & 0.869\errbar{0.008}& 0.925\errbar{0.007}\\
    BDRMIL~\cite{bdrmil}  &0.926\errbar{0.079}  &  0.905\errbar{0.092}  &  0.629\errbar{0.110} & 0.869\errbar{0.066}& 0.908\errbar{0.054}\\
    RGMIL~\cite{rgmil}  &0.940\errbar{0.070}  &  0.920\errbar{0.106}  &  0.714\errbar{0.107} & 0.842\errbar{0.088}& 0.915\errbar{0.042}\\
    TR-RGMIL~\cite{tr_rgmil}  &0.946\errbar{0.078}  &  \textbf{0.970}\errbar{0.042}  &  \textbf{0.747}\errbar{0.054} & \textbf{0.961}\errbar{0.040}& \underline{0.941}\errbar{0.054}\\
    PSMIL~\cite{psamil}  &\textbf{0.962}\errbar{0.065}  &  \underline{0.964}\errbar{0.057}  &  0.734\errbar{0.136} & 0.884\errbar{0.061}& 0.918\errbar{0.052}\\
    \rowcolor{cyan!10} Conveyance (\textbf{ours}) &\underline{0.952}\errbar{0.075}  &  0.940\errbar{0.073}  &  \underline{0.738}\errbar{0.078} & \underline{0.904}\errbar{0.070}& \textbf{0.943}\errbar{0.062}\\
    
    \bottomrule
  \end{tabular}
\end{table}

\begin{table}
\caption{Results on the Camelyon-16~\cite{camelyon_16} WSI benchmark. Values in parentheses are 95\% confidence intervals over five independent runs. \textbf{Bold}: best, \underline{underline}: second best. $^\star$ No confidence intervals provided in the original publication.}
  \label{tab:camelyon16}
  \centering \small
  \begin{tabular}{lccc}
    \toprule

  Method & ACC\,($\uparrow$) & AUC\,($\uparrow$) & F1\,($\uparrow$)\\
    \midrule
    ABMIL~\cite{abmil} &0.845\cinterval{0.839}{0.851}& 0.780\cinterval{0.769}{0.791} & 0.854\cinterval{0.848}{0.860}\\
    TransMIL~\cite{transmil} &0.858\cinterval{0.848}{0.868}& 0.797\cinterval{0.776}{0.818} & 0.906\cinterval{0.875}{0.937}\\
    DTFDMIL(AFS)~\cite{dtfdmil} &0.908\cinterval{0.892}{0.925}& 0.882\cinterval{0.861}{0.903} & \textbf{0.946}\cinterval{0.941}{0.951}\\
    DTFDMIL (MaxMinS)~\cite{dtfdmil} &0.899\cinterval{0.887}{0.912}& 0.865\cinterval{0.848}{0.882} & \underline{0.941}\cinterval{0.936}{0.946}\\
    DSMIL~\cite{dsmil} & 0.856\cinterval{0.843}{0.869}& 0.815\cinterval{0.797}{0.832} & 0.899\cinterval{0.890}{0.908}\\
    CAMIL~\cite{camil} &0.917\cinterval{0.912}{0.922}& \underline{0.959}\cinterval{0.958}{0.960} & 0.881\cinterval{0.872}{0.890}\\
    PSMIL~\cite{psamil} &\underline{0.922}$^\star$\phantom{\cinterval{0.000}{0.00}} & 0.956$^\star$\phantom{\cinterval{0.000}{0.00}} & 0.921$^\star$\phantom{\cinterval{0.000}{0.00}}\\
    \rowcolor{cyan!10}  Conveyance (\textbf{ours})  &\textbf{0.933}\cinterval{0.928}{0.939}& \textbf{0.977}\cinterval{0.977}{0.978} & 0.905\cinterval{0.898}{0.912}\\
    
    \bottomrule
  \end{tabular}
\end{table}

\subsection{Ordinal regression: age prediction}
We use three datasets for evaluation; CACD2000~\cite{cacd2000}, CLAP2016~\cite{clap2016} and UTKFace~\cite{utkface}. We compare against (i) cross-entropy and regression baselines, (ii) soft prior methods that encode ordinality through target label distributions (Soft Labels~\cite{soft_labels}, DLDL~\cite{dldl}), and (iii)~methods that introduce age-specific inductive biases via modified heads or loss terms (ORCNN~\cite{orcnn}, Mean-Variance~\cite{mean_var}, DLDL-V2~\cite{dldl_v2}). 
For \textsc{Conveyance} on CACD2000 and UTKFace, we encode a fixed ordinal neighborhood of $\pm2$ years into the transition matrix~$Q$, defining the source set as $\mathcal{S} = \{t-2,\ldots,t+2\}$ for true label~$t$ as shown in Fig.~\ref{fig:placeholder}. On CLAP2016, we evaluate an adaptive variant in which the window is set per sample to $\pm\lceil\sigma_{\text{age}}\rceil$ years, using the annotation standard deviation provided with each image as a direct measure of label ambiguity. Following the protocol established in \cite{facial_age_estimation} and consistent with existing baselines, We report Mean Absolute Error (MAE) for all methods. Additionally, for CLAP2016, we report the $\epsilon$-error~\cite{clap2016}, a metric that penalizes predictions proportionally to the annotation consensus: predictions on ambiguous images (high $\sigma$) are penalized less than predictions on images with high annotator agreement.  We use default parameters of ($\alpha{=}\beta{=}1$) for Conveyance + MAE.
Table~\ref{table:age-all} report results on three benchmarks. We discuss the two groups of baselines in turn.

\textbf{Conveyance vs.\ soft prior methods.} Among methods that encode ordinal structure through target label distributions (Table~\ref{table:age-all}), \textsc{Conveyance} achieves the best MAE on all three datasets without explicitly modeling the target distribution; instead, the loss constrains only which classes receive probability mass, leaving the data to shape the distribution within $\mathcal{S}$. 
The gain is most pronounced on CLAP2016, the noisiest benchmark, where our method reduces MAE by $0.17$ over the strongest baseline (DLDL, $5.16\to4.99$) and by $1.03$ over Cross Entropy ($6.02\to4.99$). On CACD2000 and UTKFace the margins are smaller but consistent, suggesting that encoding ordinal relationships in the loss geometry is at least as effective as smoothing target distributions. 


The right of Table~\ref{table:age-all} examines how the choice of plausible set affects performance on CLAP2016: MAE is lowest for a window of $w{=}\pm2$ and degrades as the window widens, indicating that $\mathcal{S}$ should reflect the actual ambiguity of the labels rather than being made as permissive as possible. Fig.~\ref{fig:results}B shows the joint distribution of true and predicted ages on the training data. CE fits the training ages almost exactly, concentrating its predictions on the diagonal and leaving no room for the ambiguity inherent in apparent-age labels. \textsc{Conveyance} instead spreads its predictions over neighbouring ages within the plausible window, so that the model reflects this ambiguity rather than memorizing individual annotations, consistent with its lower test MAE in Table~\ref{table:age-all}. The UMAP embeddings in Fig.~\ref{fig:results-ablation} (Appendix~\ref{sec:analysis-app}) offer a complementary geometric view: under CE they form disjoint clusters with no ordinal continuity, whereas \textsc{Conveyance} recovers a continuous curvilinear manifold that reflects the semantic progression of aging, suggesting that the performance gains reflect a better-structured representation rather than a mere loss-level artifact.

\textbf{Conveyance + MAE vs.\ ordinal methods.} We also compare against SOTA methods that exploit ordinal structure through architectural or loss-level inductive biases. Results are depicted in Table~\ref{table:age-all}. \textsc{Conveyance}~$+$~MAE matches or outperforms all baselines on every dataset. On CACD2000 and CLAP2016, it achieves the best MAE ($4.42$ and $4.63$), improving by minor margins over plain Regression and ORCNN ($4.43$ and $4.65$, respectively). On UTKFace it ties with ORCNN, a method specifically designed for ordinal regression, while requiring no bespoke ordinal architecture. Notably, Mean-Variance and DLDL-V2, despite their age-tailored formulations, underperform plain Regression on CLAP2016, highlighting that dataset noise can defeat hand-crafted priors; a regime where the Conveyance loss remains robust.

\begin{table}
\caption{ Age estimation results on CACD2000~\cite{cacd2000}, CLAP2016~\cite{clap2016}, and UTKFace~\cite{utkface}. \textbf{Left:} Comparison with two classes of methods: (i) approaches that encoding ordinal structure via soft target distributions, and (ii) methods with ordinal or age-specific inductive biases. Conveyance models ordinality through the loss geometry using a fixed $\pm2$-year window; for CLAP2016, the window is set per sample to $\pm\lceil\mathtt{std\_age}\rceil$, based on annotator disagreement. 
\textbf{Right:} Effect of the plausible set $\mathcal{S}$ (window size $w$) on CLAP2016 (mean $\pm$ std over 10 runs). Performance peaks at $w=\pm2$ (MAE $= 5.15$) and degrades for larger $w$.
}
  \label{table:age-all}
  \small
  \begin{minipage}{.6\textwidth} \centering
  \addtolength{\tabcolsep}{-0.2em}
  \begin{tabular}{lcccc}
    \toprule
      & \!\!CACD2000\!\! & \multicolumn{2}{c}{CLAP2016} & UTKFace \\
    \cmidrule(lr){2-2} \cmidrule(lr){3-4} \cmidrule(lr){5-5}
    Method & MAE & MAE & \!\!\!\!$\epsilon$-error\!\!\!\! & MAE \\
    \midrule
    \!\textit{Soft priors} \\
    ~Cross Entropy & 4.57\errbar{0.05} & 6.02\errbar{0.10} & 0.44 & 4.76\errbar{0.03} \\
    ~Soft Labels~\cite{soft_labels}   & 4.49\errbar{0.06} & 5.48\errbar{0.14} & 0.41 & 4.66\errbar{0.02} \\
    ~DLDL~\cite{dldl}          & 4.66\errbar{0.03} & 5.16\errbar{0.05} & 0.39 & 4.68\errbar{0.02} \\
    \rowcolor{cyan!10} ~Conveyance (\textbf{ours}) & \textbf{4.48}\errbar{0.04} & \textbf{4.99}\errbar{0.05} & \textbf{0.37} & \textbf{4.59}\errbar{0.03} \\
    \\[-2.5mm]
    \!\textit{Ordinal methods} \\
    ~Regression & 4.43\errbar{0.03} & 4.97\errbar{0.08} & 0.38 & 4.54\errbar{0.02} \\
    ~ORCNN~\cite{orcnn}      & 4.47\errbar{0.02} & 4.65\errbar{0.03} & 0.36 & \textbf{4.46}\errbar{0.02} \\
    ~Mean-Var~\cite{mean_var}   & 4.59\errbar{0.06} & 4.94\errbar{0.07} & 0.38 & 4.58\errbar{0.01} \\
    ~DLDL-V2~\cite{dldl_v2}    & 4.53\errbar{0.02} & 5.05\errbar{0.03} & 0.38 & 4.59\errbar{0.02} \\
    \rowcolor{cyan!10} ~Conv.\,+\,MAE (\textbf{ours}) & \textbf{4.42}\errbar{0.03} & \textbf{4.63}\errbar{0.04} & \textbf{0.35} & \textbf{4.46}\errbar{0.02} \\
    \bottomrule
  \end{tabular}
  \end{minipage}
  \hfill
  \begin{minipage}{.375\textwidth}
  \centering
    \includegraphics[width=\linewidth,height=5.5cm,keepaspectratio]{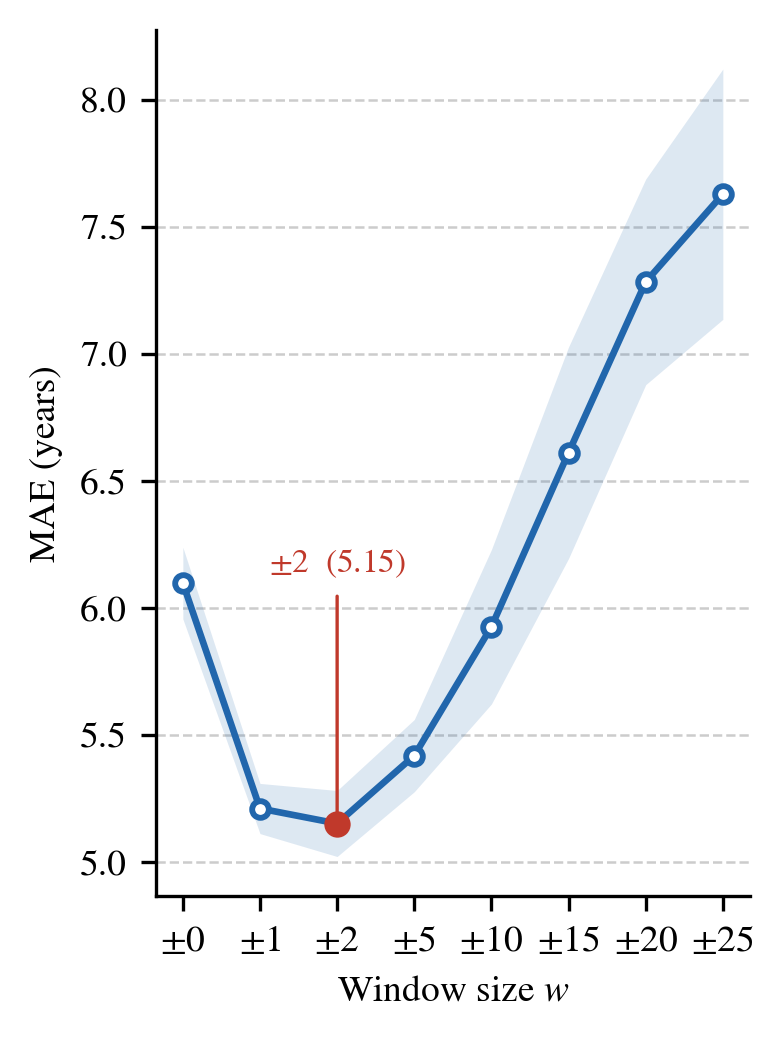}
  \end{minipage}
  
\end{table}

\subsection{Hierarchical classification: zero-shot genus classification}
Finally, we evaluate our method on CUB-200-2011~\cite{cub_dataset} using a zero-shot genus generalization test: for each of the 7 genera with $\geq4$ species, $k \in \{1,2,3\}$ randomly drawn species are held out entirely, yielding 7, 14 and 21 OOD species, respectively. Genus accuracy aggregates species probabilities by genus; predicted genus is argmax over genus-aggregated probabilities. This measures structure transfer to unseen species, and the test becomes harder as $k$ grows, since fewer congeneric species remain to define each genus during training. Each (method, $k$) pair is run with 3 seeds, and a given ($k$, seed) pair defines the same holdout set for all methods, so the reported standard deviations reflect both training noise and the choice of withheld species. All methods receive species supervision via CE; Triplet~\cite{triplet_loss}/SupCon~\cite{supcon} get additional genus structure via pair mining; \textsc{Conveyance} receives genus structure via $Q$ matrix where $Q_{i,j}=1$ for same-genus pairs (Fig.~\ref{fig:placeholder}). All trained for $100$ epochs (AdamW lr=$5\times10^{-4}$, batch=$128$), monitoring in-distribution species accuracy on the standard CUB test set. All methods use identical ViT-B/16 linear-probe (2-layer MLP: $768\to512\to(200-7k)$ dim). Baselines: CE, label smoothing ($\epsilon=0.1$), CE\,+\,Triplet ($\operatorname{margin}=0.3$), CE\,+\,SupCon ($\tau=0.07$). Conveyance ($\alpha=\beta=1$).

Table~\ref{cub-table} reports OOD genus accuracy for all five methods and all three values of $k$. In-distribution performance is broadly comparable across methods (species accuracy 84.6--85.2\%, genus accuracy 90.2--90.9\%; Appendix~\ref{sec:cub-fixed}).
The most discerning metric is therefore OOD genus accuracy, which probes whether a model's internal organization of the logit space reflects genuine taxonomic structure rather than per-species decision boundaries optimized in isolation. CE ranks last for every $k$, an expected outcome given that it imposes no constraint on inter-class relationships. Label smoothing helps at $k{=}1$ (+3.04\,pp) but its advantage over CE vanishes as $k$ grows (+0.35\,pp at $k{=}3$), as softening targets neither encodes taxonomy nor restructures the feature space. CE+Triplet does not consistently outperform label smoothing despite receiving genus information: although batch-hard mining encourages same-genus embeddings to cluster, it provides gradient signal through only one positive--negative pair per anchor per step, yielding sparse, high-variance supervision that leaves many congeneric pairs untouched. CE+SupCon is the strongest baseline precisely because it maximizes log-probability over \emph{all} same-genus pairs simultaneously, providing a richer and more stable genus-level signal. \textsc{Conveyance} attains the highest mean OOD genus accuracy for every $k$: it improves on CE+SupCon by 1.92\,pp at $k{=}1$ and 1.28\,pp at $k{=}3$, and ties with it at $k{=}2$, where the variance of all methods is dominated by a single difficult holdout split. It does so while operating entirely in logit space at $\mathcal{O}(B \cdot C)$ complexity, identical to CE.

Figure~\ref{fig:results} (panel C; $k{=}1$, see Appendix~\ref{sec:cub-fixed}) explains qualitatively, the mechanism by which our method achieves higher OOD generalization. The UMAP projections of logit-space representations show that \textsc{Conveyance} organizes congeneric species into coherent, well-separated clusters, whereas CE logits exhibit no genus-level structure, reflecting an objective that treats all classes as exchangeable. Overlaid on each projection, are the mean probability masses: $p_t$ (target species), $p_{\mathcal{S}}$ (all species of the correct genus), and $p_{\mathcal{N}}$ (all remaining species).  \textsc{Conveyance} raises $p_{\mathcal{S}}$ from 0.862 to 0.912 and suppresses $p_{\mathcal{N}}$ by 36\% relative to CE, a direct consequence of the $p_{\mathcal{N}}/p_{\mathcal{S}}$ penalty in the loss. The geometric and probabilistic evidence are mutually consistent: \textsc{Conveyance} concentrates both representational and probability mass within the taxonomically relevant set, suggesting that the $p_{\mathcal{N}}/p_{\mathcal{S}}$ penalty induces logit-space structure that transfers to unseen congeneric species.

\begin{table}[htbp]
 \caption{Zero-shot genus generalization on CUB-200-2011~\cite{cub_dataset}: genus-level top-1 accuracy (\%) on species withheld entirely from training, with $k$ species withheld from each of the 7 genera with $\geq4$ species (7, 14 and 21 OOD species for $k{=}1,2,3$). Mean\,$\pm$\,std over 3 seeds; each seed draws a different holdout set, shared by all methods. $B$, $C$, $D$ denote batch size, number of classes, and embedding dimension. In-distribution species and genus accuracy are reported in Appendix~\ref{sec:cub-fixed}.}
  \label{cub-table}
  \centering
  \small
  \begin{tabular}{llccc}
  \toprule
    & & \multicolumn{3}{c}{OOD Genus Accuracy\,($\uparrow$)} \\
  \cmidrule(lr){3-5}
  Loss Function & Complexity & $k{=}1$ & $k{=}2$ & $k{=}3$ \\
  \midrule
    CE & $\mathcal{O}(B \cdot C)$ & 77.04\errbar{1.84} & 69.05\errbar{6.11} & 56.51\errbar{0.65} \\
    Label Smoothing~\cite{label_smoothing} & $\mathcal{O}(B \cdot C)$ & 80.08\errbar{2.77} & 69.77\errbar{6.70} & 56.86\errbar{0.88} \\
    CE+Triplet Loss~\cite{triplet_loss} & $\mathcal{O}(B^2 \cdot D + B \cdot C)$\expensive{} & 79.04\errbar{1.94} & 71.37\errbar{6.22} & 57.83\errbar{1.00} \\
    CE+SupCon Loss~\cite{supcon} & $\mathcal{O}(B^2 \cdot D + B \cdot C)$\expensive{} & 80.32\errbar{2.48} & 72.89\errbar{7.37} & 59.14\errbar{0.89} \\
    \rowcolor{cyan!10} Conveyance (\textbf{ours}) & $\mathcal{O}(B \cdot C)$ & \textbf{82.24}\errbar{2.15} & \textbf{72.93}\errbar{6.60} & \textbf{60.42}\errbar{0.10} \\
  \bottomrule
\end{tabular}
\end{table}

\section{Conclusion}
\label{sec:conclusion}
We introduced \textsc{Conveyance}, a versatile framework for classification in structured class spaces.  Its knowledge encoding scheme, a boolean matrix $Q$ indicating plausible class transitions, is more tractable for practitioners than a full probability prior and strictly more expressive than any fixed parametric family.  We proved that the Conveyance loss is monotone with respect to probability mass reallocation toward the plausible set $\mathcal{S}$, and partially convex in the non-plausible logits, providing theoretical guarantees for stable optimization.  Across three task categories: label asymmetry, ordinal regression and hierarchical classification, a single objective with minimal hyperparameter tuning consistently matches or outperforms specialized baselines at the computational cost of standard cross-entropy. We aim for \textsc{Conveyance} to lower the barrier to incorporating domain structure into supervised learning across any task where class topology is known but difficult to express through generative modeling. We envision it serving as a differentiable, architecture-agnostic alternative to the cost-augmented inference required by structured output frameworks such as SSVMs.

A particularly promising direction is the extension to \emph{sample-level} transition masks, where $Q$ is instantiated per sample from auxiliary signals such as annotator disagreement scores, aleatoric uncertainty estimates, or soft pairwise similarities from a pre-trained embedding. Our adaptive $\pm\sigma$ variant on CLAP-2016 is a first step in this direction, demonstrating that per-sample ordinal windows improve both accuracy and calibration.  More broadly, a data-driven $Q$ could allow the model to effectively correct noisy or contested labels by learning the plausible neighborhood structure from the dataset itself, a capability that is structurally inaccessible to any fixed-prior objective.

\paragraph{Limitations.}\label{limitations}
First, \textsc{Conveyance} inherits its structural prior directly from $Q$. Our robustness study (Table~\ref{tab:q-robustness}) shows that moderate misspecification is tolerated and that omitting true transitions hurts more than adding spurious ones, with accuracy falling back to the CE level. This evidence is, however, limited to synthetic noise with a known support; how reliably practitioners can specify $Q$ in real annotation settings, and whether transition structure can be recovered from a partial mask, remain open questions. Second, the effective strength of the $\alpha$ term scales with the number of classes: on CIFAR-100 the same nominal $\alpha{=}0.1$ produces a ${\sim}10\times$ stronger odds penalty than on CIFAR-10, requiring per-dataset recalibration.  Whether a normalization of $\alpha$ by $|\mathcal{C}|$ eliminates this sensitivity is a hypothesis we leave for future validation.

Finally, the expressiveness of \textsc{Conveyance} is bounded by the quality of $Q$: for large-vocabulary settings such as ImageNet ($C{=}1000$) or datasets with thousands of fine-grained classes, constructing a meaningful transition mask requires structured domain knowledge, a taxonomy, ontology, or learned metric, that may not be readily available.  Automating the derivation of $Q$ from auxiliary sources such as knowledge graphs or self-supervised embeddings is a natural direction for future work.

\begin{ack}

This work was supported by institutional funding from Robert Koch Institute. The authors received no specific third-party grants or external financial support for this research

The authors declare no competing interests.
\end{ack}

\bibliographystyle{abbrv}
{
\small

\bibliography{references.bib}

}


\appendix

\clearpage

\section{Log-Sum-Exp Reformulations of the Conveyance Loss}
\label{sec:derivation}

In this appendix, we provide the derivations that connect the Conveyance loss introduced in Section \ref{sec:method} to its successive reformulations. To transform the standard Conveyance loss (Eq. \ref{eq:conveyance}) into its log-sum-exp equivalent (Eq. \ref{eq:conveyance-expanded}), we proceed as:
\begin{align*}
\ell(p,t)
&= \log \Big(
1 +
\alpha \cdot \frac{1-p_t}{p_t} +
\beta \cdot \frac{1-p_\mathcal{S}}{p_\mathcal{S}}
 \Big)\\
&= \log \Big(
\exp(0) +
\exp\Big(\log\Big(\frac{1-p_t}{p_t} \cdot \alpha\Big)\Big) +
\exp\Big(\log\Big(\frac{1-p_\mathcal{S}}{p_\mathcal{S}} \cdot \beta\Big)\Big)
 \Big)\\
&= \mathrm{LSE} \Big\{
0 \,,\,
 {\color{MidnightBlue}\log\Big(\frac{1-p_t}{p_t}\Big) + \log \alpha} \,,\,
 {\color{Mahogany}\log \Big(\frac{1-p_\mathcal{S}}{p_\mathcal{S}}\Big) + \log \beta }
\Big\}
\end{align*}
where the last expression corresponds to Eq.\ \eqref{eq:conveyance-expanded}. The terms colored in blue and brown are the two log-odds on top of which the log-sum-exp (LSE) margin structure applies. To further develop the loss to the form of Eq.\ \eqref{eq:conveyance-more}, we take these log-odds and express them in terms of their associated logits:
\begin{align*}
\color{MidnightBlue}\log\Big(\frac{1-p_t}{p_t}\Big) + \log \alpha
&= \log\Big(\frac{\sum_{c \in \mathcal{C} - \{t\}} p_c}{p_t}\Big) + \log \alpha\\
&= \log\Big(\frac{\sum_{c \in \mathcal{C} - \{t\}} \exp(z_c)}{\exp(z_t)}\Big) + \log \alpha\\
&= \log\Big(\sum_{c \in \mathcal{C} - \{t\}} \exp(z_c-z_t + \log \alpha)\Big)\\
&= \color{Green} \mathrm{LSE}\big\{ z_c - z_t + \log \alpha\big\}_{c \in \mathcal{C} - \{t\}}
\intertext{Likewise, for the second term, we get the expression:}
\color{Mahogany}\log\Big(\frac{1-p_\mathcal{S}}{p_\mathcal{S}}\Big) + \log \beta
&= \log\Big(\frac{\sum_{n \in \mathcal{N}} p_n}{\sum_{s \in \mathcal{S}} p_s}\Big) + \log \beta\\
&= \log\Big(\frac{\sum_{n \in \mathcal{N}} \exp(z_n)}{\sum_{s \in \mathcal{S}} \exp(z_s)}\Big) + \log \beta
\intertext{and defining $z_\mathcal{S} = \log\big(\sum_{s \in \mathcal{S}} \exp(z_s)\big)$, interpretable as a pseudo-logit, we continue as}
&= \log\Big(\frac{\sum_{n \in \mathcal{N}} \exp(z_n)}{\exp(z_\mathcal{S})}\Big) + \log \beta\\
&= \log\Big(\sum_{n \in \mathcal{N}} \exp\Big(z_n - z_\mathcal{S} + \log \beta\Big)\Big)\\
&= \color{Purple} \mathrm{LSE} \big\{z_n - z_\mathcal{S} + \log \beta\big\}_{n \in \mathcal{N}}
\end{align*}

Using the property that a nesting of log-sum-exp over different lists can also be expressed as a single log-sum-exp over a concatenation of these lists (defining $a$ and $b$ the lists,
$\mathrm{LSE}\{\mathrm{LSE}\{a\},\mathrm{LSE}\{b\}\} = \mathrm{LSE}\{a , b\}$%
), we get the expression of Eq.\ \eqref{eq:conveyance-more}:
\begin{align*}
\ell(p,t) = \mathrm{LSE} \Big\{ 0
\,,\,
{\color{Green}\big(z_c - z_t + \log \alpha\big)_{c \in \mathcal{C} - \{t\}}}
\,,\,
{\color{Purple}\big(z_n -  z_\mathcal{S} + \log \beta\big)_{n \in \mathcal{N}}}
 \Big\},
\end{align*}
where the outer log-sum-exp pools over $1+|\mathcal{C}-\{t\}|+|\mathcal{N}|$ elements.

\section{Algorithmic Formulation of the Conveyance Loss}
\label{sec:pseudo}

Building on LSE/logit-based formulations derived above, we provide an algorithm to evaluate the Conveyance loss, which operates directly on the logits $(z_j)_{j \in \mathcal{C}}$ and on the knowledge readily encoded in the boolean matrix $Q$. For this purpose we rely on a nested logit-based LSE formulation:
\begin{align*}
\ell(p,t) &= \mathrm{LSE} \big\{ \tau_0 \,,\,  \tau_\alpha \,,\, \tau_\beta \big\},
\end{align*}
where
\begin{align*}
\tau_0 &= 0 &
\tau_\alpha &= z_{\neg t} - z_t + \log \alpha &
\tau_\beta &= z_\mathcal{N} - z_\mathcal{S} + \log \beta,
\end{align*}
and
\begin{align*}
z_{\neg t} &= \mathrm{LSE}_{c \in \mathcal{C} - \{t\}} \{z_c\} &
z_\mathcal{N} &= \mathrm{LSE}_{n \in \mathcal{N}} \big\{ z_n \} &
z_\mathcal{S} &= \mathrm{LSE}_{s \in \mathcal{S}} \big\{ z_s \}.
\end{align*}

This nested formulation is functionally equivalent to Eqs.\ \eqref{eq:conveyance}--\eqref{eq:conveyance-more}, but combines practicality and numerical stability as it allows us to express the Conveyance loss as a sequence of logit-based calculations. A direct implementation in probability space, e.g., by computing $p_t = \operatorname{softmax}(\mathbf{z})_t$ and then evaluating Eq.~\eqref{eq:conveyance} explicitly, would be functionally equivalent, but it is numerically unstable in practice: The intermediate exponentials $\exp(z_c)$ overflow for logit magnitudes above ${\approx}88$ (with \texttt{float32}), which are possible in early training before weight norms stabilize. 
Our nested log-sum-exp formulation prevents this from materializing, as each term $\tau$ is computed as a difference of log-sum-exp scores (themselves evaluated via the standard max-subtraction trick), keeping all intermediate values in a safe numerical range regardless of logit scale. The approach is presented in Algorithm~\ref{alg:conveyance} in vectorized form, where $C$ is the number of classes and $B$ denotes the batch size, and where we leverage the masked log-sum-exp capabilities of neural network libraries such as PyTorch. This algorithm serves as a blueprint for practical implementations.

\begin{algorithm}[h]
\caption{Stable vectorized forward-evaluation of the Conveyance loss}
\label{alg:conveyance}
\KwIn{
    $\mathbf{Z} \in \mathbb{R}^{B \times C}$ (logits)\;
    $\mathbf{t} \in \mathbb{Z}^{B}$ (targets)\;
    $\mathbf{Q} \in \{0,1\}^{C \times C}$ (plausibility matrix)\;
    $\alpha, \beta \in \mathbb{R}_{>0}$ (weights)\;
}
\KwOut{Scalar loss $\mathcal{L}$}

\BlankLine
\tcp{Build plausible set $\mathcal{S}$ and non-plausible set $\mathcal{N}$}
$\mathbf{M}_{\mathcal{S}} \leftarrow \mathbf{Q}_{:,\, \mathbf{t}}^{\top}$
\tcp*{source mask from Q, shape $[B \times C]$}
$\mathbf{M}_{\mathcal{S}}[\,b,\, t_b\,] \leftarrow 1 \quad \forall\, b$
\tcp*{include target $t$ in $\mathcal{S}$}
$\mathbf{M}_{\mathcal{N}} \leftarrow \lnot\, \mathbf{M}_{\mathcal{S}}$
\tcp*{non-plausible mask}

\BlankLine
\tcp{Extract relevant log-sum-exp scores}
$\mathbf{z}_t \leftarrow \mathbf{Z}[\,b,\, t_b\,] \quad \forall\, b$
\tcp*{target logit, shape $[B]$}
$\mathbf{z}_{\mathcal{S}} \leftarrow \operatorname{LSE}(\mathbf{Z},\; \mathbf{M}_{\mathcal{S}})$
\tcp*{$\log \sum_{s \in \mathcal{S}} \exp(z_s)$, shape $[B]$}
$\mathbf{z}_{\mathcal{N}} \leftarrow \operatorname{LSE}(\mathbf{Z},\; \mathbf{M}_{\mathcal{N}})$
\tcp*{$\log \sum_{n \in \mathcal{N}} \exp(z_n)$, shape $[B]$}
$\mathbf{z}_{\lnot t} \leftarrow \operatorname{LSE}(\mathbf{Z},\; \lnot\{t\})$
\tcp*{$\log \sum_{c \neq t} \exp(z_c)$, shape $[B]$}

\BlankLine
\tcp{Compute the three LSE terms}
$\tau_0 \leftarrow \mathbf{0}$
\tcp*{the constant ``1'' in $\log(1 + \cdots)$}
$\tau_\alpha \leftarrow \log \alpha + \mathbf{z}_{\lnot t} - \mathbf{z}_t$
\tcp*{$\log\, \alpha \cdot \frac{1 - p_t}{p_t}$}
$\tau_\beta  \leftarrow \log \beta  + \mathbf{z}_{\mathcal{N}} - \mathbf{z}_{\mathcal{S}}$
\tcp*{$\log\, \beta \cdot \frac{p_{\mathcal{N}}}{p_{\mathcal{S}}}$}

\BlankLine
\tcp{Soft-max pool over terms and reduce}
$\boldsymbol{\ell} \leftarrow \operatorname{LSE}(\tau_0,\, \tau_\alpha,\, \tau_\beta)$
\tcp*{shape $[B]$}
$\mathcal{L} \leftarrow \operatorname{reduce}(\boldsymbol{\ell})$
\tcp*{mean or sum over batch}

\Return{$\mathcal{L}$}
\end{algorithm}

\section{Comprehensive Experimental Details}
\label{sec:exp-details}

\subsection{Datasets}
\label{sec:datasets}
\paragraph{CIFAR-10 and CIFAR-100 \cite{cifar10-100}}
CIFAR-10 and CIFAR-100 are standard benchmarks for image classification, both consisting of 60{,}000 color images of resolution $32{\times}32$ pixels, split into 50{,}000 training samples and 10{,}000 test samples. CIFAR-10 contains 10 mutually exclusive classes (airplane, automobile, bird, cat, deer, dog, frog, horse, ship, truck), each with 6{,}000 images (5{,}000 for training, 1{,}000 for testing). 

CIFAR-100 contains 100 fine-grained classes grouped into 20 coarser superclasses, with 5 subclasses per superclass; each fine-grained class has 500 training and 100 test images. Both datasets are well-balanced across classes. For CIFAR-10 we use standard data augmentation (random cropping with padding 4 and random horizontal flipping), while for CIFAR-100 we additionally apply random rotation. Test images are evaluated without augmentation, using only per-channel normalization with the dataset statistics.

\textbf{CACD2000}~\cite{cacd2000} is a large-scale celebrity face dataset collected from the web, containing 163{,}446 images of 2000 identities with ages ranging from 14 to 62 (mean 38). The benchmark uses a single fixed train/val/test split partitioned at the identity level (1{,}800/80/120 identities; approximately 145{,}000/7{,}600/10{,}600 images), preventing identity leakage across sets. Because labels are derived by matching celebrity names to birth years in web images, label noise from mis-dated or mis-attributed photographs is expected.

\textbf{UTKFace}~\cite{utkface} contains 24107 images spanning a wide age range (1 to 116, mean 33) across five racial groups. It is evaluated under 5-fold cross-validation over ten equal folds (6 train / 2 validation / 2 test per fold), yielding approximately 14{,}460/4{,}820/4{,}820 images per split. Identities are not tracked, so split assignment is image-level.

\textbf{CLAP2016}~\cite{clap2016} is the training partition of the ChaLearn Looking at People 2016 apparent-age challenge, containing 7{,}591 images with ages ranging from 1 to 96 (mean 31). Unlike the other datasets, labels reflect crowd-sourced apparent age rather than chronological age: each image carries an annotation standard deviation (\texttt{std\_age}) with a mean of 4.25 years and a maximum of 14.12 years, and 32.8\% of images have $\mathtt{std\_age} > 5$ years, making this the noisiest of the three benchmarks. The original challenge split is retained: 4{,}113 training / 1{,}500 validation / 1{,}978 test images.

\paragraph{CUB-200-2011. \cite{cub_dataset}}
The Caltech-UCSD Birds-200-2011 dataset is a fine-grained visual classification benchmark consisting of 11{,}788 images of 200 North American bird species, split into 5{,}994 training and 5{,}794 test images. Species are organized into a two-level taxonomic hierarchy spanning 123 genera and 27 families. Each image is annotated with a bounding box, 15 part locations, and 312 binary attributes; we use only the bounding box annotations to crop images to the bird region before resizing. The dataset presents significant intra-class variation (pose, lighting, occlusion) and high inter-class similarity, particularly among congeneric species, making it a standard benchmark for fine-grained recognition.For our zero-shot genus generalization protocol, we further partition the training set by withholding $k \in \{1,2,3\}$ species per genus from each of the 7 genera containing at least four species, yielding 7, 14 or 21 held-out species ($200-7 \times k$ training classes) used exclusively as an OOD probe set at evaluation time.

\paragraph{Camelyon-16 \cite{camelyon_16}.}
A whole-slide image (WSI) benchmark for binary tumor detection in sentinel lymph node sections, released as part of the ISBI 2016 Grand Challenge. The dataset comprises 400 gigapixel H\&E-stained slides (270 training, 130 test), each labeled as tumor-positive or tumor-negative at the slide level. In the MIL formulation, each WSI is a bag and non-overlapping $256{\times}256$ patches extracted at $20\times$ magnification are instances; patch-level labels are not used during training. Following~\cite{dsmil}, patch features are extracted with a ResNet-50 pretrained via SimCLR on the training slides.

\paragraph{MUSK1 and MUSK2 \cite{musks}.}
Two molecule-activity benchmarks for drug discovery originally proposed by ~\cite{original_musk}.
Each bag is a molecule and each instance is a low-energy conformation of that molecule; a bag is labeled positive (musk) if at least one conformation is responsible for a musky odor. MUSK1 contains 92 bags (47 positive) and 476 instances with 166 real-valued features. MUSK2 is a larger and harder variant with 102 bags (39 positive) and 6{,}598 instances sharing the same feature space.

\paragraph{Elephant, Fox, and Tiger \cite{animals}.}
Three image-based MIL benchmarks drawn from the Animals dataset.
Each dataset contains 200 bags (100 positive, 100 negative), where each bag is a natural image of the target animal or a distractor animal, and instances are image regions described by 230-dimensional color and texture histogram features.
The task is binary: predict whether the image contains the target animal. Fox is widely regarded as the most challenging of the three due to the high intra-class variation and frequent occlusion of the animal.

\subsection{Noise Types}
\label{sec:noise-types}
Here we illustrate and explain each of the four noise structures we considered.

\begin{figure}[h]
  \centering
  \includegraphics[width=\linewidth]{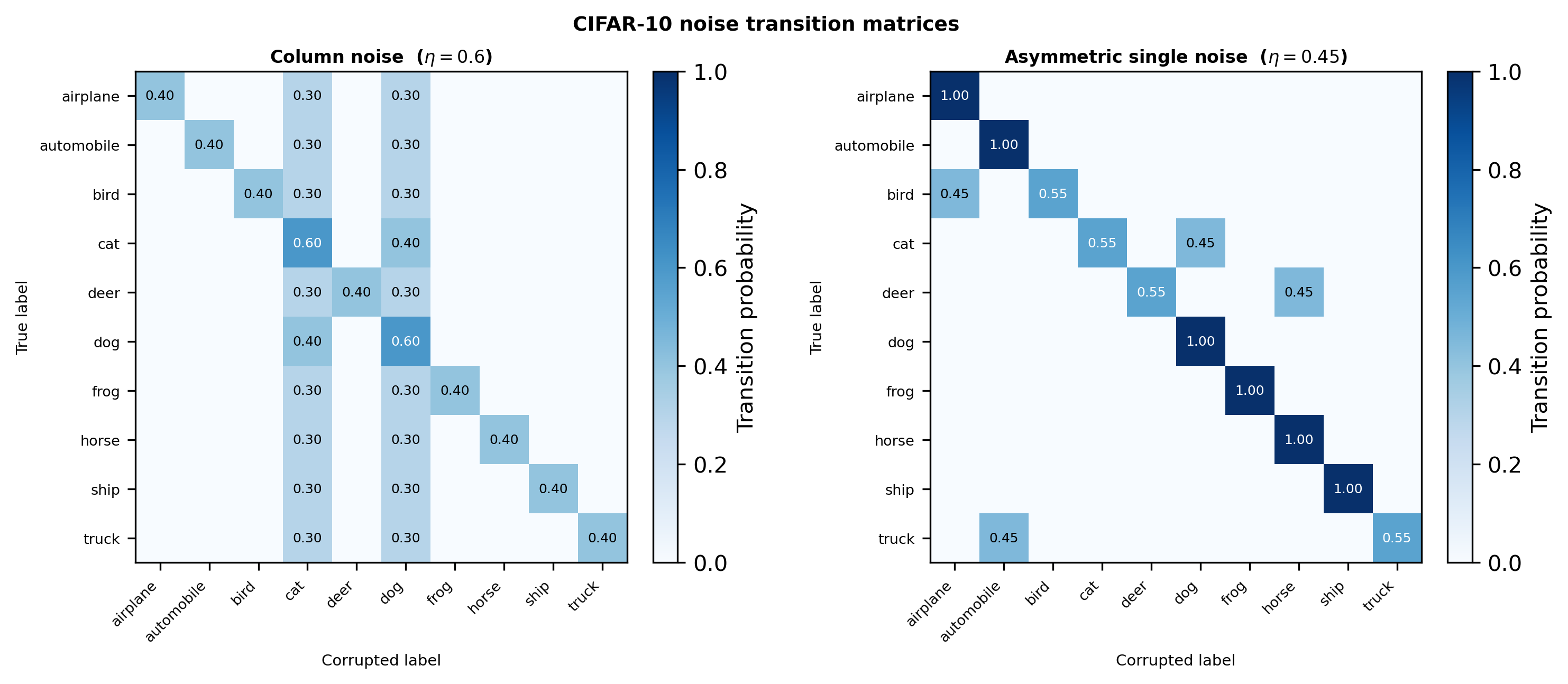}
  \caption{%
    Transition matrices $T$ for the two CIFAR-10 noise types used in our experiments.
    Each entry $T_{ij}$ gives the probability that a true-class-$i$ instance receives
    corrupted label $j$.
    \emph{Left (Column noise ($\eta{=}0.6$))}:Every class loses $60\%$ of its labels to the global sinks cat (class~3) and dog (class~5), making them label-space attractors while all other classes are unaffected as destinations.
    \emph{Right (Asymmetric single noise ($\eta{=}0.45$))}: Each class is corrupted along a single directed edge (e.g., cat$\to$dog, automobile$\to$truck), yielding a near-diagonal structure with isolated off-diagonal entries. Transition probabilities are annotated in each cell.
  }
  \label{fig:noise-structure-cifar10}
\end{figure}

\begin{figure}[h]
  \centering
  \includegraphics[width=\linewidth]{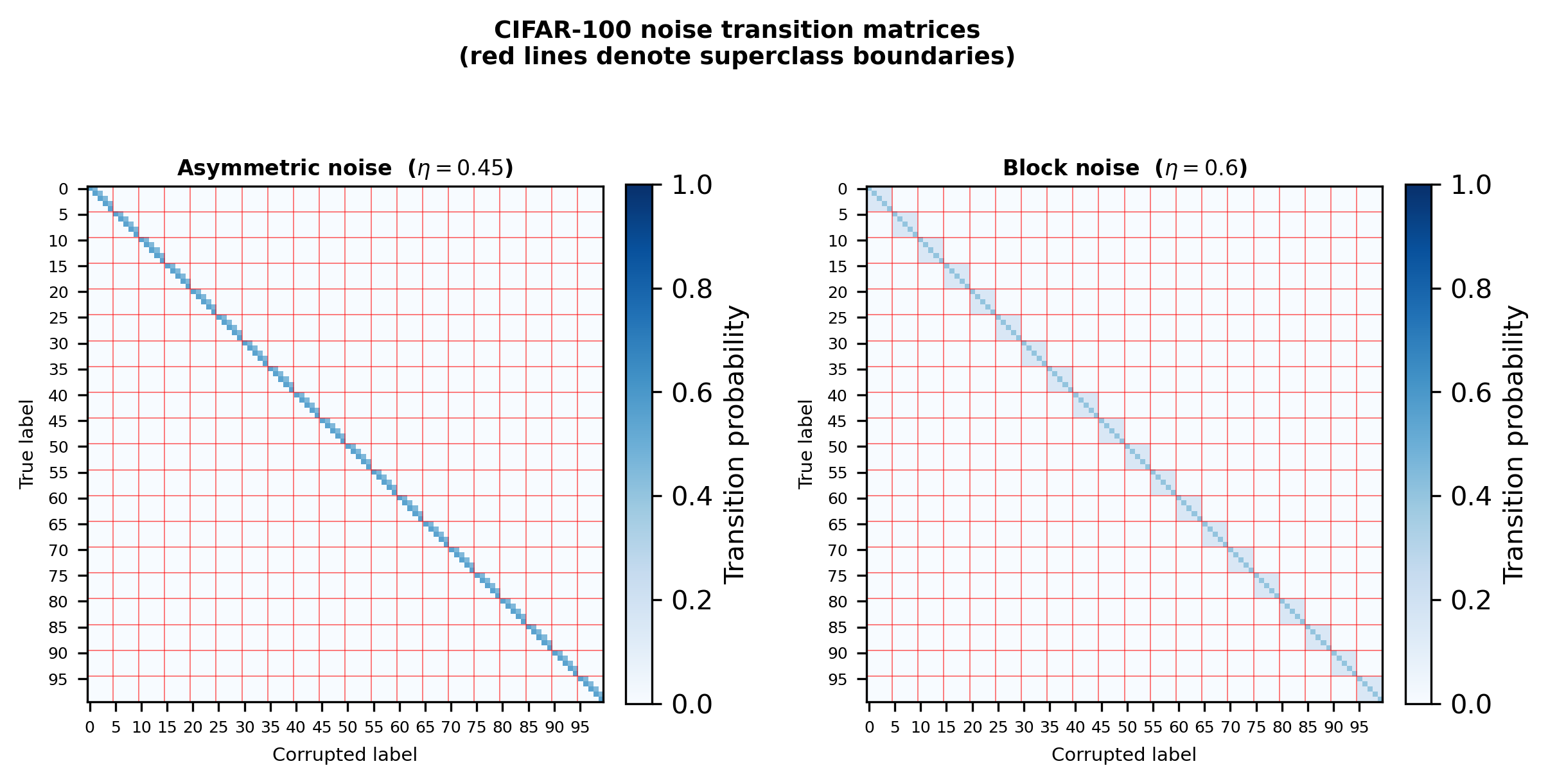}
  \caption{%
    Transition matrices $T$ for the two CIFAR-100 noise types used in our experiments. Red lines delineate the 20 superclass boundaries (5 fine classes each).
    \emph{Left (Asymmetric noise ($\eta{=}0.45$))}: Corruption follows a single directed cyclic chain \emph{within} each superclass, producing a sparse near-diagonal structure whose off-diagonal mass is entirely confined inside the $5{\times}5$ superclass blocks.
    \emph{Right (Block noise ($\eta{=}0.6$))}:  Within each superclass block, every fine class distributes $60\%$ of its labels uniformly across the other four co-superclass members, resulting in dense $5{\times}5$ blocks of equal off-diagonal probability ($0.15$ each) aligned with superclass boundaries.
  }
  \label{fig:noise-structure-cifar100}
\end{figure}
\paragraph{CIFAR-10: Column noise ($\eta{=}0.6$).} A depiction of column noise on CIFAR-10 is shown in Figure ~\ref{fig:noise-structure-cifar10}.
Column noise concentrates all corruption into two fixed destination classes: cat (class~3)
and dog (class~5).
Every source class $c \notin \{3, 5\}$ retains $40\%$ of its labels and distributes the
remaining $60\%$ equally between cat and dog ($30\%$ each).
The destination classes themselves are partially self-corrupting: cat retains $60\%$ of
its own labels and sends the remaining $40\%$ to dog, and symmetrically dog retains $60\%$
and sends $40\%$ to cat.
Formally, the transition matrix $T$ satisfies:
\begin{align*}
    T_{c,c} = 0.4,\quad T_{c,3} = T_{c,5} = 0.3 \quad &\forall\; c \notin \{3,5\}, \\
    T_{3,3} = 0.6,\quad T_{3,5} = 0.4, \quad
    T_{5,5} = 0.6,\quad T_{5,3} = 0.4.
\end{align*}
The result is a heavily imbalanced noise structure in which cat and dog receive mislabeled
samples from all other eight classes simultaneously, while the remaining classes only lose
samples.

\paragraph{CIFAR-10: Asymmetric noise ($\eta{=}0.45$).} A depiction of asymmetric noise on CIFAR-10 is shown in Figure ~\ref{fig:noise-structure-cifar10}.
Asymmetric noise applies four independent one-directional label flips between
semantically similar class pairs, each with flip rate $45\%$:
\begin{align*}
    \text{truck} &\to \text{automobile}, &
    \text{bird}  &\to \text{airplane}, \\
    \text{cat}   &\to \text{dog},       &
    \text{deer}  &\to \text{horse}.
\end{align*}
Each source class retains $55\%$ of its labels and sends $45\%$ to its paired destination.
The four destination classes (automobile, airplane, dog, horse) receive additional
mislabeled samples exclusively from their single paired source.
All other classes are unaffected ($T_{c,c}{=}1$).

\paragraph{CIFAR-100: Asymmetric noise ($\eta{=}0.45$).}
A depiction of asymmetric noise on CIFAR-100 is shown in Figure ~\ref{fig:noise-structure-cifar100}.
Asymmetric noise applies the same cyclic one-directional flip scheme independently within
each superclass: subclass $i$ sends $45\%$ of its labels to subclass $i{+}1$
(mod~5), retaining $55\%$.
The last subclass in each group sends to the first, completing the cycle.
This results in $100$ active noise edges in total, all confined within superclass
boundaries, with every class simultaneously acting as a source and a destination.
The corruption is thus fully intra-superclass, leaving inter-superclass label assignments
unperturbed.

\paragraph{CIFAR-100: Block noise ($\eta{=}0.6$).}
A depiction of block noise on CIFAR-100 is shown in Figure ~\ref{fig:noise-structure-cifar10}.
Block noise applies uniform within-superclass corruption.
For every class $c$ within a superclass of 5 members, $60\%$ of labels are corrupted and
distributed \emph{uniformly} among the four co-superclass members ($15\%$ each), while
$40\%$ of labels are retained:
\begin{equation*}
    T_{c,c} = 0.4, \qquad
    T_{c,c'} = 0.15 \quad \forall\; c' \neq c \text{ in the same superclass}.
\end{equation*}
All inter-superclass entries are zero.
Unlike asymmetric noise, block noise introduces no preferred direction of corruption:
every pair of co-superclass members can confuse each other symmetrically, creating a dense
$5{\times}5$ uniform confusion block for each of the 20 superclasses.

\subsection{Implementation details}
\label{sec:implementation-details}
Following, we illustrate the benchmarks, baselines and implementation details for each use case.
For synthetic label noise, we follow ~\cite{masking}, by introducing column noise and block noise for CIFAR-10 and CIFAR-100 respectively. Additionally, we test out asymmetric noise on both datasets.  Similarly to Masking~\cite{masking}, our method receives a \emph{binary} transition matrix encoding the noise structure where each entry indicates whether a given class can act as a source of corruption for another. Formally, the model is provided with the source set $\mathcal{S}_c = \{c : Q_{ct} > 0\}$ for each class $c$, where $Q$ is the true (unknown) transition matrix. The prediction accuracy is used to evaluate the classification performance of each
model in the test set. Following \cite{APL, AFL, ANL, drainage}, we use an 8-layer CNN for CIFAR-10 and a ResNet-34 \cite{resnet} for CIFAR-100. We adopt the code base provided by ~\cite{ANL}. Models are trained with batch size 128, for 120 and 200 epochs on CIFAR-10, and CIFAR-100, respectively, using SGD with momentum 0.9, cosine learning rate decay, and initial learning rates of 0.01. For both models on all noise settings for Conveyance, $\alpha{=}1/\beta$, $\beta=10$ (we refer the reader to Appendix ~\ref{sec:ablations}). For Masking~\cite{masking}, we use the same reconstructor backbone and batch size as the other baselines, and follow the original training procedure~\cite{masking} for the generator and discriminator. We tune the weight decay over $\{0, 10^{-5}, 10^{-4}, 10^{-3}, 10^{-2}\}$ and report the best result for each noise type. 

We evaluate Conveyance on \textbf{Multiple Instance Learning} (MIL), where bag-level supervision is the only training signal and patch-level labels are unavailable. We consider two regimes: the Camelyon-16~\cite{camelyon_16} whole-slide image (WSI) tumor-detection benchmark, and five classic tabular MIL benchmarks (Elephant~\cite{animals}, Fox~\cite{animals}, Tiger~\cite{animals}, Musk1~\cite{musks}, Musk2~\cite{musks}). For both benchmarks, we use the publicly available codebases of~\cite{psamil,dsmil} for evaluation (MIT License). For Camelyon-16 we compare against ABMIL~\cite{abmil}, TransMIL~\cite{transmil}, DTFD-MIL (AFS and MaxMinS)~\cite{dtfdmil}, DSMIL~\cite{dsmil}, CAMIL~\cite{camil}, and PSMIL~\cite{psamil}. Following~\cite{dsmil,psamil}, we report accuracy (ACC), AUC, and macro F1-score with 95\% confidence intervals over five independent runs; each run performs stratified 5-fold cross-validation on the 270 official training bags, aggregating per-fold predictions on the 129 fixed held-out test bags. Following~\cite{dsmil,psamil}, Camelyon-16 patch features are extracted with a ResNet-50 encoder pretrained via SimCLR. The model (a two-layer gated-attention network) is trained with AdamW ($\text{lr}{=}10^{-3}$, weight decay~$10^{-4}$, dropout~$0.1$) for 40 epochs while using default parameters in the Conveyance loss ($\alpha{=}\beta{=}1$). For the classic benchmarks, we follow the 10-fold cross-validation protocol of~\cite{psamil}, running five independent repetitions and reporting mean accuracy with 95\% confidence intervals. Following~\cite{psamil,dsmil}, we use a two-layer MLP backbone (hidden dim 512) with a gated-attention head, trained for 40 epochs with $\text{lr}{=}10^{-3}$ and weight decay~$10^{-4}$ while setting $\alpha{=}\beta{=}1$ in the Conveyance loss. The asymmetric noise structure is encoded for both benchmarks in the transition matrix $Q$, where $Q_{\text{neg}\to\text{pos}}{=}1$ and $Q_{\text{pos}\to\text{neg}}{=}0$, reflecting the natural asymmetry of MIL as shown in Fig.\ \ref{fig:placeholder}.

We evaluate Conveyance on the \textbf{age estimation} problem. Experiments are conducted on three benchmarks: CACD2000~\cite{cacd2000}, CLAP2016~\cite{clap2016}, and UTKFace~\cite{utkface}. We compare against two natural baselines; cross-entropy and MAE. We compare against two groups of methods. The first group encodes ordinal structure by replacing the one-hot target with a soft label distribution: Soft Labels~\cite{soft_labels} use a Laplace-like decay over ordinal distance, and DLDL~\cite{dldl} places a fixed-$\sigma$ Gaussian centered on the true label. The second group goes further by either modifying the prediction head or adding loss-level inductive biases: ORCNN~\cite{orcnn} replaces the standard $K$-class head with $K{-}1$ binary ordinal classifiers; Mean-Variance~\cite{mean_var} augments cross-entropy with explicit moment regularization on the predicted age distribution; and DLDL-V2~\cite{dldl_v2} extends DLDL with an additional $\ell_1$ regression term on the predicted mean age.. Performance is reported in terms of Mean Absolute Error (MAE). Following~\cite{facial_age_estimation}, we adopt their  curated dataset splits (train/val/test), designed to avoid inflated performance estimates so that improvements can be attributed to the method rather than favorable data partitioning. We run all methods using the codebase provided by ~\cite{facial_age_estimation}, which is under MIT License.  All models use a ResNet-50 backbone pretrained on ImageNet-1K with $256\times256$ input resolution, trained for 50 epochs with the AdamW optimizer ($\text{lr}{=}10^{-4}$) and a batch size of 100. As depicted in Fig.\ \ref{fig:placeholder}, for ordinal regression, we use a window of $\pm2$ years (i.e.\ the source set is $\mathcal{S} = \{t-2,\ldots,t+2\}$ for true label~$t$), and use default parameters for $\alpha,\beta$ in Conveyance~$+$~MAE across all datasets ($\alpha{=}\beta{=}1$). All remaining hyperparameters for all methods follow the settings of~\cite{facial_age_estimation} on the \emph{same} benchmarks. Additionally, we evaluate a $\sigma$-window variant on CLAP2016, where the per-sample window is set to $\pm\lceil\mathtt{std\_age}\rceil$ years using the crowd-sourced annotation standard deviation provided with each image, encoding per-sample label uncertainty directly into the source set $\mathcal{S}$.

For \textbf{hierarchical classification}, we evaluate Conveyance on CUB-200-2011, a fine-grained bird classification benchmark comprising 200 species organized into 123 genera. To assess whether the loss encodes genus-level structure beyond the training distribution, we construct a zero-shot genus generalization test: for each of the 7 genera with at least four species, $k \in \{1,2,3\}$ species are withheld entirely from training, yielding 7, 14 or 21 held-out species. The withheld species are drawn per seed (3 seeds per $k$), and a given ($k$, seed) pair yields the same holdout set for all methods. At test time, we measure genus-level top-1 accuracy on these unseen species; a model succeeds only if it assigns the correct genus to images of species it has never encountered. In addition, we report species-level and genus-level top-1 accuracy on the standard CUB held-out test set to measure in-distribution performance (Appendix~\ref{sec:cub-fixed}). All methods share the same ViT-B/16~\cite{vit} backbone in linear-probe mode, with a two-layer MLP head (Linear$\to$BN$\to$ReLU$\to$Linear, hidden dim 512). Models are trained for 100 epochs with AdamW (\text{lr}=$5\times10^{-4}$). We compare Conveyance against four baselines: cross-entropy (CE), label smoothing (LS)~\cite{label_smoothing}, CE with batch-hard triplet loss (CE+Triplet)~\cite{triplet_loss}, and CE with supervised contrastive loss (CE+SupCon)~\cite{supcon}. Triplet and SupCon operate on the 512-dimensional hidden representation and use genus-level groupings to define positive pairs. For Conveyance, the source set $S$ consists of all species belonging to the same genus as the target class (Fig.~\ref{fig:placeholder}). Concretely, the transition matrix $Q$ is set to $Q_{c,t}=1$ when species $c$ and $t$ share a genus and $Q_{c,t}=0$ otherwise. We sweep label-smoothing $\epsilon \in {0.05,0.1,0.15,0.2}$, triplet margin $\in {0.1,0.2,0.3,0.4,0.5}$, and SupCon temperature $\tau \in {0.05,0.07,0.09,0.11}$ while monitoring species accuracy. For Conveyance, we set $\alpha=\beta=1$ and consistent with our ablations in Appendix~\ref{sec:ablations}, due to the large number of classes ($C=200-7 \times k$), we lower the learning rate to $1\times10^{-4}$.

\subsection{Computational resources.}
\label{sec:compute}
All experiments were conducted on a single NVIDIA H100 GPU (80GB VRAM) with an Intel Xeon CPU and 64GB system memory on an internal HPC cluster. On CIFAR-10 and CIFAR-100, individual training runs required approximately 20 and 35 minutes, respectively. For the multiple instance learning (MIL) benchmarks~\cite{animals,musks}, each run required approximately 2 hours, while experiments on Camelyon16~\cite{camelyon_16} took around 4 hours per run. For the age estimation datasets, each method required approximately 8 hours of compute, and for the CUB dataset~\cite{cub_dataset}, around 1 hour per method. In total, the experiments reported in this paper required approximately 300 GPU hours. Additional compute was used for preliminary and failed experiments.

\section{Ablations}
\label{sec:ablations}
\subsection{Annotation bias}
\begin{figure}[h]
  \centering
\includegraphics[width=1\textwidth]{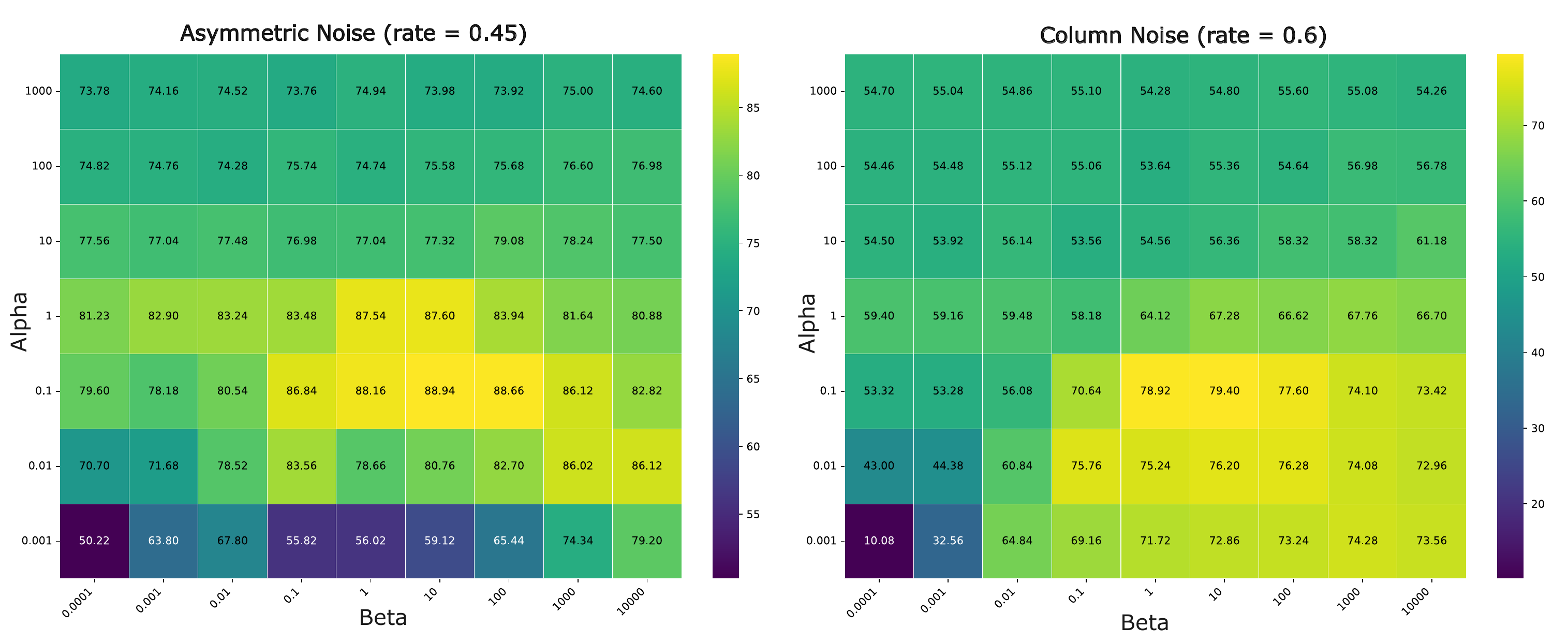}
    \caption{Validation accuracy (\%) as a function of $\alpha$ and $\beta$ on CIFAR-10 under asymmetric single noise ($\eta{=}0.45$, left) and column noise ($\eta{=}0.6$, right). Under asymmetric noise, high accuracy is sustained across a broad plateau ($\alpha \approx 0.1$, $\beta \in [1, 1000]$), reflecting the recoverable binary per-pair confusion structure. Under column noise, the optimal region is sharper ($\alpha \approx 0.1$, $\beta \in [1, 100]$): large $\alpha$ inadvertently reinforces noisy destination labels (cat, dog), while small $\beta$ fails to compress source-class predictions. The default $( \alpha{=}0.1 ,\, \beta{=}10)$ lies within the optimal region for both noise types.}
      \label{fig:ablation-heatmap-cifar10}

\medskip

\includegraphics[width=1\textwidth]{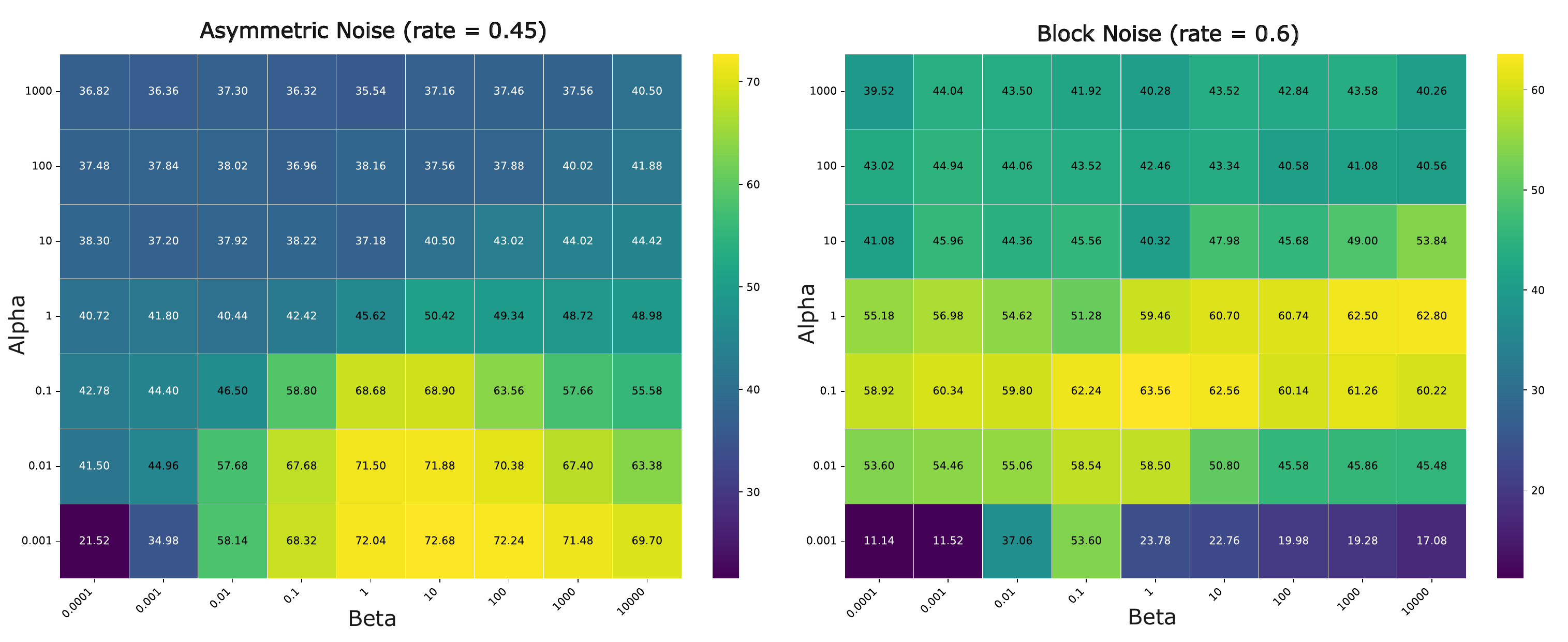}
  \caption{Validation accuracy (\%) as a function of $\alpha$ and $\beta$ on CIFAR-100 under
asymmetric noise ($\eta{=}0.45$, left) and block noise ($\eta{=}0.6$, right). Under asymmetric noise, the optimal $\alpha$ shifts to $10^{-3}$--$10^{-2}$ relative to CIFAR-10, compensating for the larger class count inflating the magnitude of the odds term; the best configuration $(\alpha{=}10^{-3} ,\,\beta{=}10)$ achieves $72.68\%$, surpassing the clean CE baseline of $69.42\%$. Under block noise, the landscape is flatter in $\beta$ but requires $\alpha \in [0.1,1]$ to break within-superclass symmetry; collapse to ${\approx}20\%$ occurs when $\beta$ is large but $\alpha$ is too small to discriminate among the four co-superclass sources. The default $(\alpha{=}0.1 ,\, \beta{=}10)$ remains competitive across both settings.}
  \label{fig:ablation-heatmap-cifar100}

\end{figure}

We conduct a grid search over $\alpha, \beta \in \{10^{-4}, 10^{-3}, 10^{-2}, 10^{-1}, 1, 10, 10^2, 10^3, 10^4\}$ on CIFAR-10 under column ($\eta{=}0.6$)  and asymmetric single ($\eta{=}0.45$) noise, and on CIFAR-100 under asymmetric ($\eta{=}0.45$) and block ($\eta{=}0.6$) noise. Appendix~\ref{sec:noise-types} explains each noise in detail. Results are summarized in Figure~\ref{fig:ablation-heatmap-cifar10} and ~\ref{fig:ablation-heatmap-cifar100} respectively. Recall that $\alpha$ governs the standard odds term, while $\beta$ controls the penalty on probability mass outside the source-augmented set.

The optimal behavior of $\alpha$ and $\beta$ is better understood by examining how each term acts on the two structurally distinct sample types defined by the noise. We partition classes into \emph{destination} classes (those receiving mislabeled samples from other classes) and \emph{source} classes ( those that lose a fraction of their samples to destination classes). As we show below, $\alpha$ and $\beta$ play fundamentally different and complementary roles for these two types, and the optimal operating point reflects the relative difficulty of each role under a given noise structure.

\textbf{CIFAR-10 Column noise:} For a sample assigned a \emph{destination} noisy label under column noise (e.g.\ noisy label~$=$~cat), all nine other classes are sources, so $\mathcal{S}$ spans the entire class set and $p_N \equiv 0$: the $\beta$ term vanishes entirely and the loss reduces to $\log(1 + \alpha(1-p_t)/p_t)$.

A large $\alpha$ therefore strongly drives the model toward the noisy destination label even for originally clean samples, explaining the collapse to the noisy-CE baseline
(${\approx}55\%$) whenever $\alpha \geq 100$. Conversely, for a \emph{source}, $\mathcal{S} = \{t\}$
and the loss becomes $\log(1 + (\alpha{+}\beta)(1-p_t)/p_t)$: both $\alpha$ and $\beta$ add constructively toward the correct class. This asymmetry defines the optimal operating point: keeping $\alpha$ small (so that the push toward destination/noisy labels remains weak) while raising $\beta$ (to sharpen source-class predictions) yields the best performance. The peak at $(\alpha{=}0.1 ,\, \beta{=}10)$, $79.40\%$, precisely reflects this trade-off.

\textbf{CIFAR-10 Asymmetric noise:} Each destination class (e.g.\ automobile) has exactly one source (truck), giving $\mathcal{S} = \{t, \mathrm{source}\}$, a small two-element set. The $\beta$ term is therefore active for destination samples: it penalizes probability mass outside the correct pair ($p_N/(p_t{+}p_{\mathrm{src}})$), continuously pushing the model to concentrate confusion within the true (clean, noisy) pair. A robust model that learns to predict the clean label satisfies this penalty naturally ($p_{\mathrm{src}} \approx 1 \Rightarrow p_N \approx 0$), while $\alpha$ only weakly anchors the noisy label. Because $\beta$ actively organizes mass into the correct pair for \emph{any} sufficiently large value, the optimal plateau is wide: $\alpha \approx 0.1$, $\beta \in [1, 1000]$, with the peak at $(\alpha{=}0.1 ,\, \beta{=}10)$ at $88.94\%$. The broader optimum relative to column noise reflects the fact that binary per-pair confusion is far easier to recover than the all-class funneling imposed by column noise.

\textbf{CIFAR-100 Asymmetric noise.}
CIFAR-100 asymmetric noise follows a cyclic chain within each of the 20 superclasses (each subclass corrupts $45\%$ of labels to the next), making every class simultaneously a source and a destination with exactly one incoming source. The source set is therefore always a two-element pair, $\mathcal{S}= \{t,\, t_{\mathrm{prev}}\}$, structurally identical to the CIFAR-10 asymmetric case with far more noise transitions. However, the optimal $\alpha$ shifts : from $\approx 0.1$ on CIFAR-10 to $0.001$-$0.01$ on CIFAR-100 (best: $\alpha{=}10^{-3} ,\, \beta{=}10$, $72.68\%$). This is not coincidental. The $\alpha$ term computes $\alpha(1-p_t)/p_t$, where $(1-p_t)$ aggregates probability mass over all competing classes. With 100 classes instead of 10, the softmax distribution is naturally more diffuse and even a well-performing model achieves a lower $p_t$ per class, making $(1-p_t)$ intrinsically larger; an $\alpha{=}0.1$ appropriate for CIFAR-10 therefore produces a $10\times$ stronger effective penalty on CIFAR-100, over-suppressing confidence in the correct class even when the prediction is accurate. Reducing $\alpha$ by the same factor restores the correct operating scale. 

As on CIFAR-10, performance degrades monotonically as $\alpha$ increases beyond $0.1$, collapsing to near noisy-CE Baseline (${\approx}38\%$) for $\alpha \geq 10$ regardless of $\beta$, while very small $\beta \leq 10^{-3}$ is also insufficient to organize mass into the correct pair ($21$-$44\%$). The optimal $\beta$ range of $[1, 100]$ mirrors the CIFAR-10 asymmetric result, confirming that the source-set packing mechanism is scale-invariant with respect to class count, while $\alpha$ must be re-calibrated.

\textbf{CIFAR-100 Block noise.}
Block noise introduces a qualitatively different interaction between $\alpha$ and $\beta$. Each class has four sources  (all other members of the same superclass) so $\mathcal{S}$ covers the entire superclass (all 5 subclasses), and $p_N$ aggregates the probability mass of the remaining 95 out-of-superclass classes. The $\beta$ term therefore acts as a \emph{superclass-level packing} force, continuously penalizing any probability mass that escapes the correct superclass. The $\alpha$ term, acting on the full $(1-p_t)/p_t$ ratio, must then provide the \emph{within-superclass discrimination} signal to break the 5-way symmetry among co-superclass members. This two-level structure creates a super-class collapse visible in the heatmap: for large $\beta \geq 10$ combined with very small $\alpha{=}10^{-3}$, performance hovers between $17.0\%$ and $22.0\%$. The is a result of $\beta$ successfully concentrating all probability mass within the correct superclass, while $\alpha{=}10^{-3}$ providing no discriminative pressure among the 5 co-superclass members. The optimal region therefore requires balancing both roles: $\alpha \in [0.1, 1]$ is needed to break within-superclass symmetry, while $\beta \in [0.1, 10^4]$ is broadly sufficient once $\alpha$ is in range (best: $\alpha{=}1 ,\, \beta{=}10$, $63.56\%$). The resulting performance landscape is flat over a much wider $\beta$ range than for asymmetric noise, consistent with the superclass-packing role of $\beta$ being achievable across many orders of magnitude once the within-superclass competition is properly regulated by $\alpha$.

\paragraph{Summary and default configuration.}
Across all four settings, the combination $(\alpha{=}0.1,\, \beta{=}10)$ used as the default in the main experiments lies within or immediately adjacent to the optimal region in every heatmap, confirming that it constitutes a robust, single choice that requires neither per-dataset nor per-noise-type tuning. A particularly striking result emerges for CIFAR-100 asymmetric noise: by reducing $\alpha$ to $10^{-3}$ to compensate for the larger class count, Conveyance achieves $72.68\%$ , surpassing the clean CE baseline of $69.42\%$ by $+3.26$ percentage points. This counter-intuitive gain suggests that the source-set grouping induced by $\beta$ acts as an implicit structural regularizer: by constraining probability mass within the correct within-superclass pair, it suppresses spurious confidence toward visually similar classes in other superclasses,  yielding better generalization even relative to a noise-free training signal.

\section{Additional Analysis}
\label{sec:analysis-app}

\begin{figure}[h]
  \centering
  \includegraphics[width=1\textwidth,clip,trim=0 0 0 15]{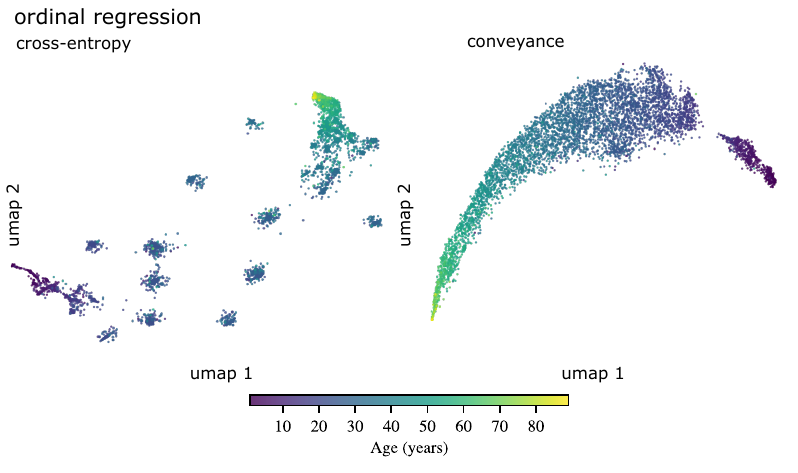}
    \caption{UMAP embeddings for CLAP2016's train and test set representation by CE and Conveyance, colored by age. CE creates fragmented clusters while overfitting to younger ages. Conveyance organizes the latent space into a smooth age manifold.}
  \label{fig:results-ablation}

\end{figure}

\subsection{Hierarchical classification: in-distribution metrics}
\label{sec:cub-fixed}
Table~\ref{tab:cub-fixed} complements Table~\ref{cub-table} with in-distribution metrics. It reports the single-split variant of the experiment ($k{=}1$, the same 7 species withheld in every run, 5 seeds), which is also the setting visualized in Fig.~\ref{fig:results}C. In-distribution species and genus accuracy are broadly comparable across methods, so the methods differ mainly in how well genus structure transfers to unseen species. The OOD column differs from the $k{=}1$ column of Table~\ref{cub-table} because the latter draws a different set of withheld species for each seed.

\begin{table}[h]
 \caption{CUB-200-2011~\cite{cub_dataset}, fixed split ($k{=}1$). Species and Genus Accuracy are on the held-out test set (193 in-distribution classes); OOD Genus Accuracy is genus-level top-1 on the 7 fully withheld species. Mean\,$\pm$\,std are reported over 5 seeds.}
  \label{tab:cub-fixed}
  \centering
  \small
  \begin{tabular}{lccc}
  \toprule
  Loss Function 
    & \makecell[c]{Species\\Accuracy\,($\uparrow$)}
    & \makecell[c]{Genus\\Accuracy\,($\uparrow$)}
    & \makecell[c]{OOD Genus\\Accuracy\,($\uparrow$)}
    \\
  \midrule
    CE &84.7\errbar{0.1}&90.2\errbar{0.2}&78.0\errbar{1.1} \\
    Label Smoothing~\cite{label_smoothing} &84.6\errbar{0.0}&90.2\errbar{0.3}&81.2\errbar{1.4} \\
    CE+Triplet Loss~\cite{triplet_loss} &85.0\errbar{0.1}&90.4\errbar{0.2}&80.9\errbar{1.0}  \\
    CE+SupCon Loss~\cite{supcon} &\textbf{85.2}\errbar{0.1}&90.8\errbar{0.2}&84.1\errbar{0.7}\\
    \rowcolor{cyan!10} Conveyance (\textbf{ours}) &84.9\errbar{0.1}&\textbf{90.9}\errbar{0.3}&\textbf{85.1}\errbar{0.5} \\
  \bottomrule
\end{tabular}
\end{table}
\clearpage

\end{document}